\documentclass[10pt,journal,compsoc]{IEEEtran}

\ifCLASSOPTIONcompsoc
  \usepackage[nocompress]{cite}
\else
  \usepackage{cite}
\fi

\ifCLASSINFOpdf
\else
\fi

\usepackage{times}
\usepackage{epsfig}
\usepackage{graphicx}
\usepackage{amsmath}
\usepackage{amssymb}
\usepackage{subcaption}
\usepackage{overpic}
\usepackage{multicol}
\usepackage{multirow}
\usepackage{bm}
\usepackage{balance}
\usepackage{booktabs}
\usepackage{color}
\usepackage{ulem}
\usepackage{hyperref}
\usepackage{ulem}

\begin{document}
\title{DREAM\raisebox{0.12\baselineskip}{+}: Efficient Dataset Distillation by Bidirectional Representative Matching}

\author{Yanqing Liu\textsuperscript{*},
        Jianyang Gu\textsuperscript{*},
        Kai Wang\textsuperscript{†},
        Zheng Zhu,
        Kaipeng Zhang,
        Wei Jiang,
        and~Yang You\textsuperscript{‡}
\IEEEcompsocitemizethanks{\IEEEcompsocthanksitem Y. Liu, K. Wang, Y. You are with the National University of Singapore, Singapore. (Corresponding author: Yang You. e-mail: youy@comp.nus.edu.sg)
\IEEEcompsocthanksitem J. Gu and W. Jiang are with Zhejiang University, China.
\IEEEcompsocthanksitem Z. Zhu is with Tsinghua University, China.
\IEEEcompsocthanksitem K. Zhang is with Shanghai AI Laboratory, China.
}%

\thanks{\textsuperscript{*}Equal contribution, \textsuperscript{†}Project lead} 
}

\markboth{IEEE TRANSACTIONS ON PATTERN ANALYSIS AND MACHINE INTELLIGENCE}%
{Shell \MakeLowercase{\textit{et al.}}: Bare Demo of IEEEtran.cls for Computer Society Journals}

\IEEEtitleabstractindextext{%
\begin{abstract}
Dataset distillation plays a crucial role in creating compact datasets with similar training performance compared with original large-scale ones. This is essential for addressing the challenges of data storage and training costs. Prevalent methods facilitate knowledge transfer by matching the gradients, embedding distributions, or training trajectories of synthetic images with those of the sampled original images. Although there are various matching objectives, currently the strategy for selecting original images is limited to naive random sampling. We argue that random sampling overlooks the evenness of the selected sample distribution, which may result in noisy or biased matching targets. Besides, the sample diversity is also not constrained by random sampling. Additionally, current methods predominantly focus on single-dimensional matching, where information is not fully utilized. To address these challenges, we propose a novel matching strategy called Dataset \textbf{D}istillation by \uline{Bidirectional} \textbf{RE}present\textbf{A}tive \textbf{M}atching (DREAM\raisebox{0.09\baselineskip}{+}), which selects representative original images for bidirectional matching. DREAM\raisebox{0.09\baselineskip}{+} is applicable to a variety of mainstream dataset distillation frameworks and significantly reduces the number of distillation iterations by more than 15 times without affecting performance. Given sufficient training time, DREAM\raisebox{0.09\baselineskip}{+} can further improve the performance and achieve state-of-the-art results. We have released the code at \href{https://github.com/lyq312318224/DREAM/}{github.com/NUS-HPC-AI-Lab/DREAM\raisebox{0.09\baselineskip}{+}}.
\end{abstract}

}

\maketitle

\IEEEdisplaynontitleabstractindextext

\IEEEpeerreviewmaketitle

\IEEEraisesectionheading{\section{Introduction}\label{sec:introduction}}

\IEEEPARstart{T}{he} development of deep learning has ushered in a remarkable era of achievements in computer vision, as evidenced by numerous influential works~\cite{he2016deep,dosovitskiy2020image,redmon2016you,liu2021swin,tsai2018learning,goodfellow2020generative,danelljan2017eco,zheng2023preventing}. 
However, these achievements are often established upon massive datasets, especially for recent large-scale models~\cite{chiangvicuna,touvron2023llama,zhu2023minigpt,hu2023bliva}.
In addition to the extraordinary effort for data collection and processing, the dependency on massive data, in turn, leads to severe problems for common deep learning practices~\cite{wang2018dataset,paul2021deep,toneva2018empirical}.
On the one hand, training on such large datasets requires enormous calculation resources, which can be infeasible for resource-restricted researchers. 
On the other hand, the storage and maintenance demands for massive data are also hard to afford. 
~\cite{zhao2020dc,he2022masked}. In response, various methodologies have emerged to tackle the cumbersome data burden by compressing the scale of the training data~\cite{wang2018dataset,sorscher2022beyond,qin2023infobatch,daquan2022lossless}.

A group of methods attempt to address the problem through selecting representative samples from the original dataset, denoted as coreset methods~\cite{lapedriza2013all,toneva2018empirical}. 
However, along with the selection, a large number of samples are directly deserted, where certain information for encapsulating the full essence of the dataset is lost. 
As a result, the performance is often not satisfactory under high compression ratios~\cite{iyer2021submodular,killamsetty2021glister,killamsetty2021grad}.
On the other hand, dataset distillation has emerged as a leading strategy, aiming to distill the information of the whole dataset into surrogate sets of manageable sizes~\cite{mtt,wang2022cafe,idc,cui2022dc,ftd}. 
This paradigm begins with a small number of learnable image tensors and iteratively refines them through alignment with various facets of the original data, including training gradients~\cite{zhao2020dc,idc}, embedding distributions~\cite{dm,wang2022cafe}, or training trajectories~\cite{mtt,ftd}.
This noble pursuit has become pivotal in addressing the data problem and has attracted significant scholarly attention~\cite{geng2023survey,sachdeva2023data,lei2023comprehensive,yu2023dataset}.

Despite the notable performance gains and compression ratios achieved by dataset distillation~\cite{idc}, a persistent challenge remains, which is the prolonged duration of the distillation process. For example, for distilling information into 50 images per class (IPC) on CIFAR-10 dataset, the expert trajectory training time and distillation time for MTT~\cite{mtt} take approximately 16 hours or more. IDC~\cite{idc} requires more than 20 hours of distillation time to achieve considerable performance. On CIFAR-100 dataset, IDC requires more than 40 hours to finish 50 IPC distillation.
We claim that the training efficiency of dataset distillation methods is largely influenced by two key elements: the sampling strategy for selecting original matching images and the optimization objectives.

Dataset distillation enriches the information in synthetic images by aligning training characteristics~\cite{zhao2020dc,dm}. Normally, random sampling is adopted for forming a mini-batch of original images to reduce the required memory in the training stage~\cite{zhao2020dc,dm}. However, random sampling often overlooks the evenness of sample distribution, for which those with larger training gradients may dominate the optimization process for gradient matching~\cite{wang2022cafe}. Besides, random sampling also fails to constrain the diversity within small batches, leading to unfaithful representation of the original data.
Another aspect overlooked by previous dataset distillation methods is the optimization objectives. 
With various alignment paradigms proposed~\cite{idc,zhao2020dc,zhao2021dsa,dm}, there are not yet works attempting to fuse these optimization targets together. 
We argue that the single-dimensional optimization objective cannot thoroughly reflect the characteristics of the original data, and hence restricts the training efficiency. 

\begin{figure}
\centering
\begin{subfigure}{0.48\textwidth}
\includegraphics[width=\textwidth]{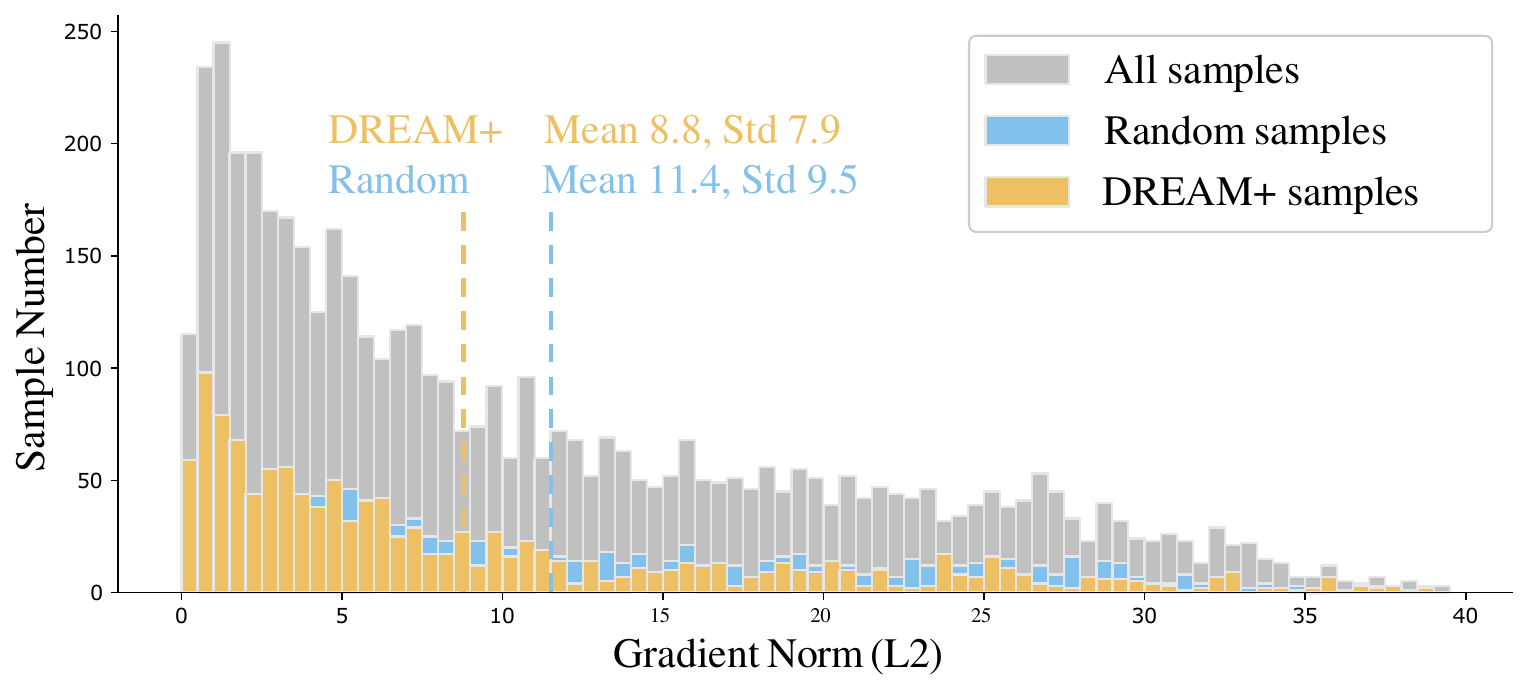}
\caption{The gradient norm distribution of the ship class in CIFAR-10. }
\label{fig:gradient}
\end{subfigure}
\begin{subfigure}{0.5\textwidth}
    \includegraphics[width=\textwidth]{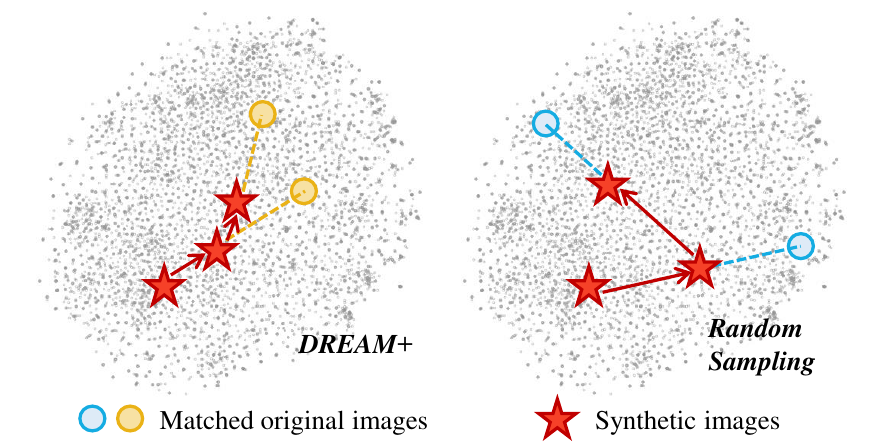}
    \caption{The migration of synthetic samples during training. }
    \label{fig:intro}
\end{subfigure}
\caption{Samples on the decision boundaries usually provide larger gradients, which biases the gradient matching optimization. Random sampling (left) overlooks the evenness of of the selected sample distribution, resulting in unstable optimization process of the synthesized samples. By only matching with proper gradients from representative original samples, our proposed DREAM\raisebox{0.09\baselineskip}{+} (right) greatly improves the training efficiency of dataset distillation tasks. Best viewed in color. }
\end{figure}

Accordingly, we introduce a novel method named as Efficient Dataset \textbf{D}istillation by \uline{Bidirectional} \textbf{RE}present\textbf{A}tive \textbf{M}atching (DREAM\raisebox{0.09\baselineskip}{+}) for more efficient dataset distillation. 
First, for sample selection, a clustering process is performed periodically within each class to generate sub-clusters that reflect the sample distribution.
The samples closest to the center of each sub-cluster are selected to form the target mini-batch for matching.
Selection of such center samples serves the dual purpose of representing nearby samples and achieving uniform coverage of the entire class distribution.
As illustrated in Fig.~\ref{fig:gradient}, the clustering-based selection results in a set of samples with less gradient variance compared with random sampling.
Secondly, we integrate both forward features and backward gradients to provide bidirectional optimization directions. 
This bidirectional matching paradigm significantly improves the training stability, leading to a smoother and more robust distillation process. For the synthetic image initialization, we adopt a clustering-based strategy, akin to~\cite{cui2022dc}, where center samples from each sub-cluster are employed.

DREAM\raisebox{0.09\baselineskip}{+} can be easily integrated into current dataset distillation frameworks. 
Comparative evaluations against common random sample selection and single-dimensional matching techniques highlight DREAM\raisebox{0.09\baselineskip}{+}'s ability to enhance the training efficiency. 
We conduct extensive experiments to demonstrate that DREAM\raisebox{0.09\baselineskip}{+} achieves comparable performance to baseline methods in less than one-fifteenth the number of iterations required.
Moreover, with the same training iteration set as other state-of-the-art methods, DREAM\raisebox{0.09\baselineskip}{+} achieves even better performance.
For example, DREAM\raisebox{0.09\baselineskip}{+} surpasses IDC by 2.3\% on CIFAR-100 with 10 IPC. 

This work expands upon our earlier conference paper~\cite{liu2023dream} and introduces several new contributions:
\begin{itemize}
    \item DREAM\raisebox{0.09\baselineskip}{+}, an enhanced version of DREAM, effectively addresses the training efficiency issue associated with single-dimensional matching during dataset distillation. The improved matching technique better captures the characteristics of the original data. 
    \item The experiments across diverse datasets and dataset distillation techniques demonstrates that DREAM\raisebox{0.09\baselineskip}{+} further accelerates training by over 15 times without compromising the distillation performance.
    \item Beyond the core methodology, we provide supplementary results, analyses, and visualizations that delve into the intricacies of bidirectional matching, offering a more comprehensive understanding of this innovative component.
\end{itemize}

\section{Related Works}
\subsection{Coreset Selection}
Coreset selection selects a subset of data based on specific metrics~\cite{guo2022deepcore,coleman2019selection}. Lapedriza et al. measure sample importance based on the benefits gained from model training on each sample~\cite{lapedriza2013all}. Toneva et al. observe that samples exhibit varying forgetting characteristics, with easily forgettable samples containing more information~\cite{toneva2018empirical}. Coreset-based methods are also widely used in continual learning~\cite{rebuffi2017icarl,aljundi2019gradient,wiewel2021condensed} and active learning tasks~\cite{sener2017active}. Shleifer et al. expedite neural network architecture search by selecting a group of "easier" samples~\cite{shleifer2019using}. While coreset-based methods are practical, they face limitations in extracting rich information from a small subset of original samples, restricting their ability to further enhance compression ratios.

\subsection{Dataset distillation} 
Dataset distillation is implemented by synthesizing image samples guided by various optimization objectives. Wang et al. introduce the concept of dataset distillation from the perspective of optimization and update synthetic images using a meta-learning approach~\cite{wang2018dataset}. Subsequent works employ a variety of optimization targets to constrain image synthesis, including matching training gradients~\cite{zhao2020dc,zhao2021dsa,jiang2022delving}, embedding distributions~\cite{dm,wang2022cafe}, and training trajectories~\cite{mtt} of original images. IDC injects additional information into synthetic samples under fixed storage constraints~\cite{idc}. IDM optimizes distribution matching by expanding feature dimensions and model parameter space~\cite{zhao2023improved}. Nguyen et al. develop a distributed meta-learning framework and incorporate kernel approximation methods~\cite{nguyen2021dataset}. RFAD accelerates the metric computation through random feature approximation~\cite{loo2022efficient}. HaBa leverages data hallucination networks to construct base images and enhance the representation capability of distilled datasets~\cite{liu2022dataset}. FRePo introduces an efficient meta-gradient computation method and a ``model pool" to mitigate the overfitting towards specific architectures~\cite{zhou2022dataset}. Some methods use generative models to complete dataset distillation, such as GLaD~\cite{cazenavette2023generalizing} and ITGAN~\cite{zhao2022synthesizing}, which compress datasets into latent variables in feature space and then use decoders for data synthesis. DiM~\cite{wang2023dim} transfers knowledge by distilling datasets into generative models.

\begin{figure*}[t]
\centering
\small
\begin{subfigure}{0.32\textwidth}
    \includegraphics[width=\textwidth]{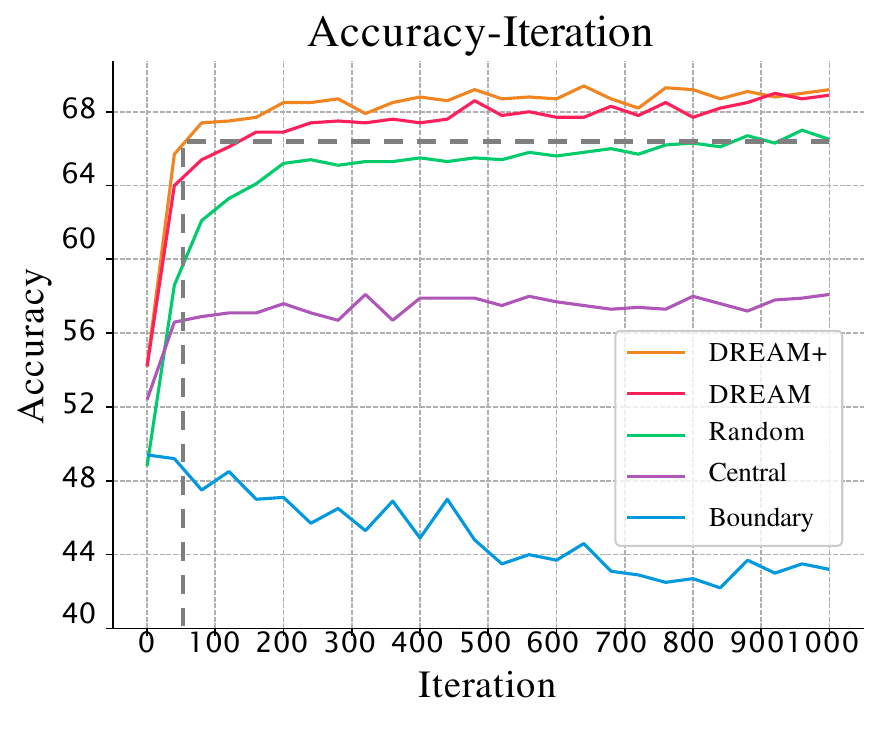}
    \caption{The accuracy curve with different strategies for selecting original images.}
    \label{fig:acc}
\end{subfigure}
\hfill
\begin{subfigure}{0.32\textwidth}
    \includegraphics[width=\textwidth]{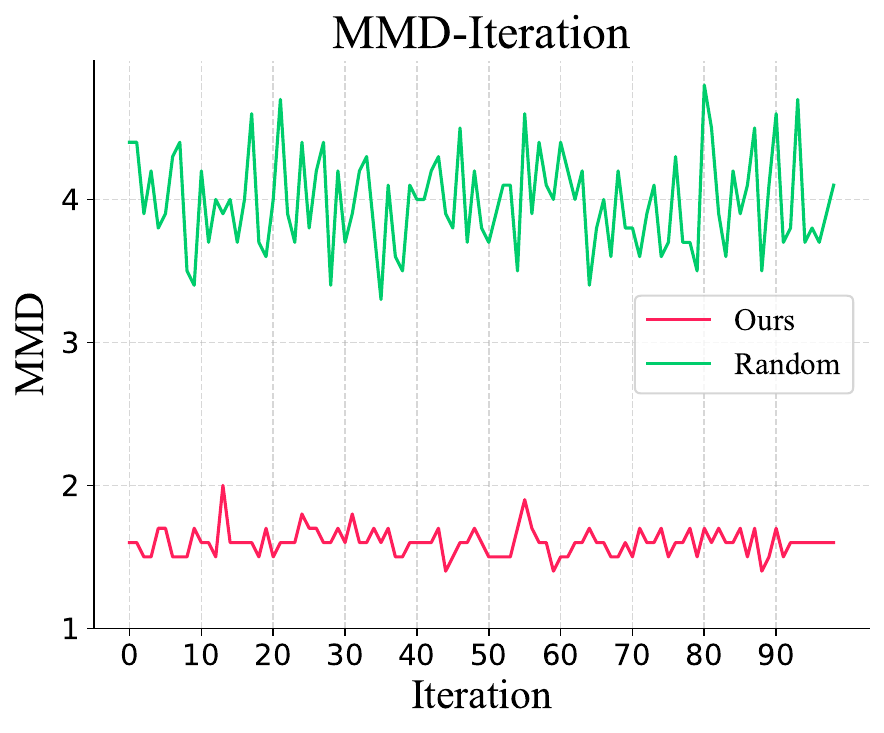}
    \caption{The MMD curve between the sampled mini-batch and the corresponding class data.}
    \label{fig:mmd}
\end{subfigure}
\hfill
\begin{subfigure}{0.32\textwidth}
    \includegraphics[width=\textwidth]{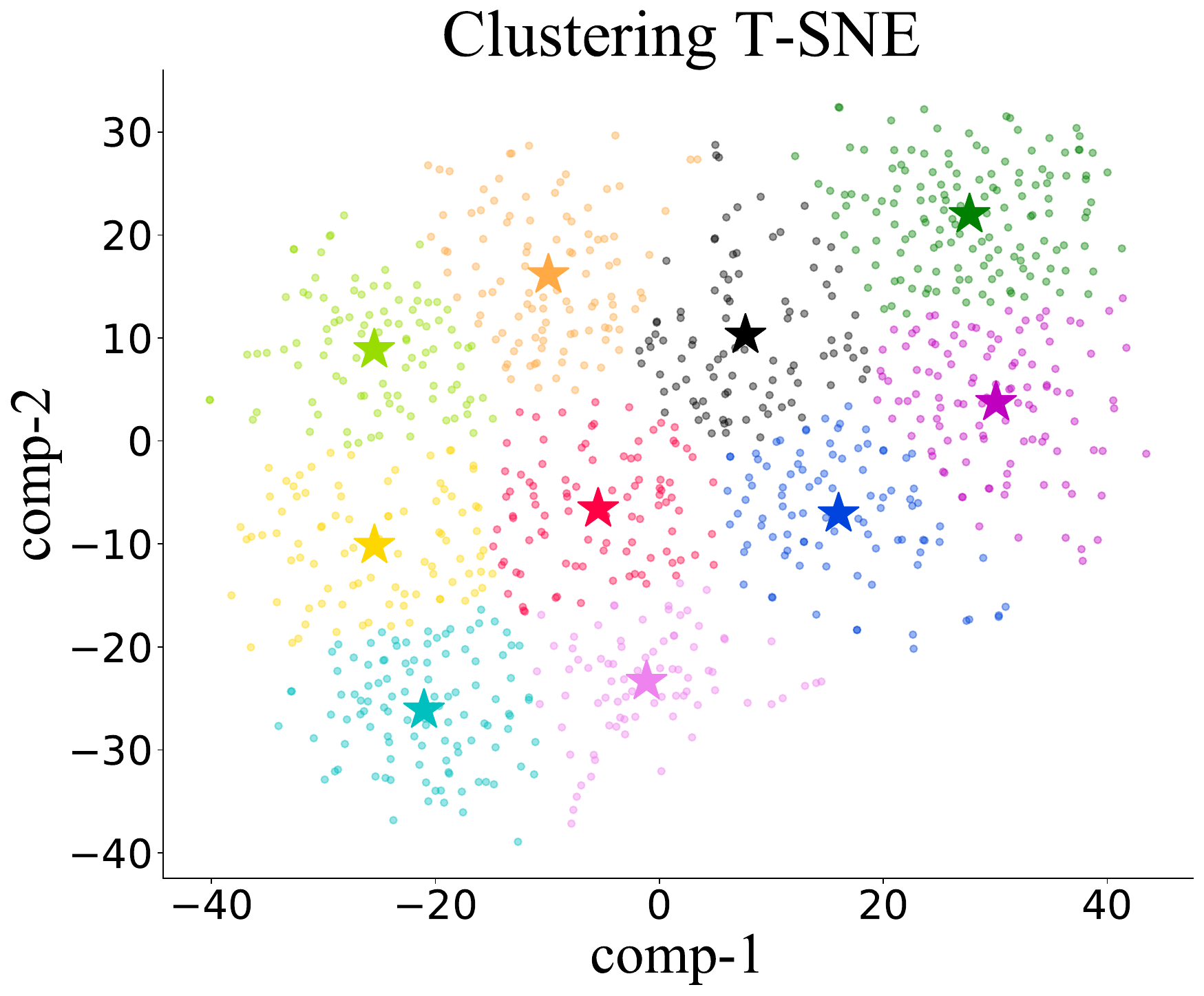}
    \caption{Example clustering and sub-cluster center results of DREAM\raisebox{0.09\baselineskip}{+}. }
    \label{fig:cluster}
\end{subfigure}
\caption{The original images obtained by random sampling have uneven distributions, which may result in noisy or biased matching targets. Besides, the coverage of random sampling on the whole sample space is low and has large fluctuations during training. Comparatively, the centers selected by DREAM\raisebox{0.09\baselineskip}{+} (stars) are representative for corresponding sub-clusters, and are evenly distributed over the whole class feature space. Experiments for (a) and (b) are conducted under 10 images-per-class setting on CIFAR-10. Best viewed in color. }
\label{fig:method}
\end{figure*}

Dataset distillation methods significantly enhance compression ratios by incorporating more information into synthetic images. However, recent state-of-the-art methods often require a large number of iterations to achieve desired validation accuracy, indicating low training efficiency. In this work, we focus on designing a novel matching strategy to improve the efficiency of dataset distillation training.

\subsection{Clustering}

Clustering is an unsupervised technique used to group data samples into distinct clusters~\cite{rehioui2016denclue}. Several clustering methods exist, each with its unique characteristics and applications.
K-means~\cite{k-means,k-means++} is a well-known method that requires specifying the number of target clusters. It optimizes the data partition to create clusters with similar sizes~\cite{intra-distance}.
Density-Based Spatial Clustering of Applications with Noise (DBSCAN) relies on density and does not necessitate a prior knowledge of the number of clusters. It gradually forms clusters by including data points within a specified tolerance range~\cite{dbscan}. DBSCAN is versatile and can handle datasets of various shapes. However, it has some limitations, including unstable cluster sizes, exclusion of outliers from clusters, and potential merging of closely located clusters.
Hierarchical clustering methods encompass two main approaches: Agglomerative and Divisive. The former progressively merges multiple clusters until a predefined condition is met, resulting in a hierarchical structure. Conversely, the latter divides a cluster into smaller segments, iteratively refining the hierarchy~\cite{hierarchical}.

\subsection{Differences from Related Works}

Several recent works have been proposed to improve the efficiency of dataset distillation. It's essential to understand the distinctions between these approaches and our proposed method.
Random Feature Approximation for Dataset Distillation (RFAD) reduces the computational complexity associated with Kernel Inducing Points (KIP) by employing random feature approximation~\cite{loo2022efficient}. RFAD primarily targets computational complexity reduction within the context of KIP. In contrast, our proposed method, DREAM\raisebox{0.09\baselineskip}{+}, concentrates on improving the training efficiency by introducing bidirectional matching strategies with selected representative original images. There are no contradictory between these two approaches. Instead, they address different aspects of efficiency issues for dataset distillation.

Jiang et al. analyze the limitations of the gradient matching method and introduce the concept of matching multi-level gradients~\cite{jiang2022delving}. Additionally, there are other methods such as those by Lorraine et al.~\cite{lorraine2020optimizing} and Vicol et al.~\cite{vicol2022implicit}, which examine shortcomings in existing techniques from the perspective of two-level optimization and enhance efficiency accordingly.
In contrast, DREAM\raisebox{0.09\baselineskip}{+} addresses training efficiency from the perspective of sampling and matching objectives within optimization-based methods. It offers seamless integration with various dataset distillation approaches, resulting in a substantial reduction in required training iterations.
These distinctions emphasize the unique contributions of DREAM\raisebox{0.09\baselineskip}{+} in the area of dataset distillation efficiency.

\section{Method}

Aiming at tackling the issue of low training efficiency in dataset distillation tasks, we propose a novel distillation approach denoted as Dataset \textbf{D}istillation by \uline{Bidirectional} \textbf{RE}present\textbf{A}tive \textbf{M}atching (DREAM\raisebox{0.09\baselineskip}{+}). DREAM\raisebox{0.09\baselineskip}{+} is designed to enhance the stability and robustness of the training process by focusing on bidirectional matching with representative original images. In this section, we outline the foundational training framework for dataset distillation, share our analysis on the training efficiency problem, and provide a comprehensive overview of the DREAM\raisebox{0.09\baselineskip}{+} methodology.

\begin{figure*}[t]
\centering
\small
\centering
    \includegraphics[width=0.9\textwidth]{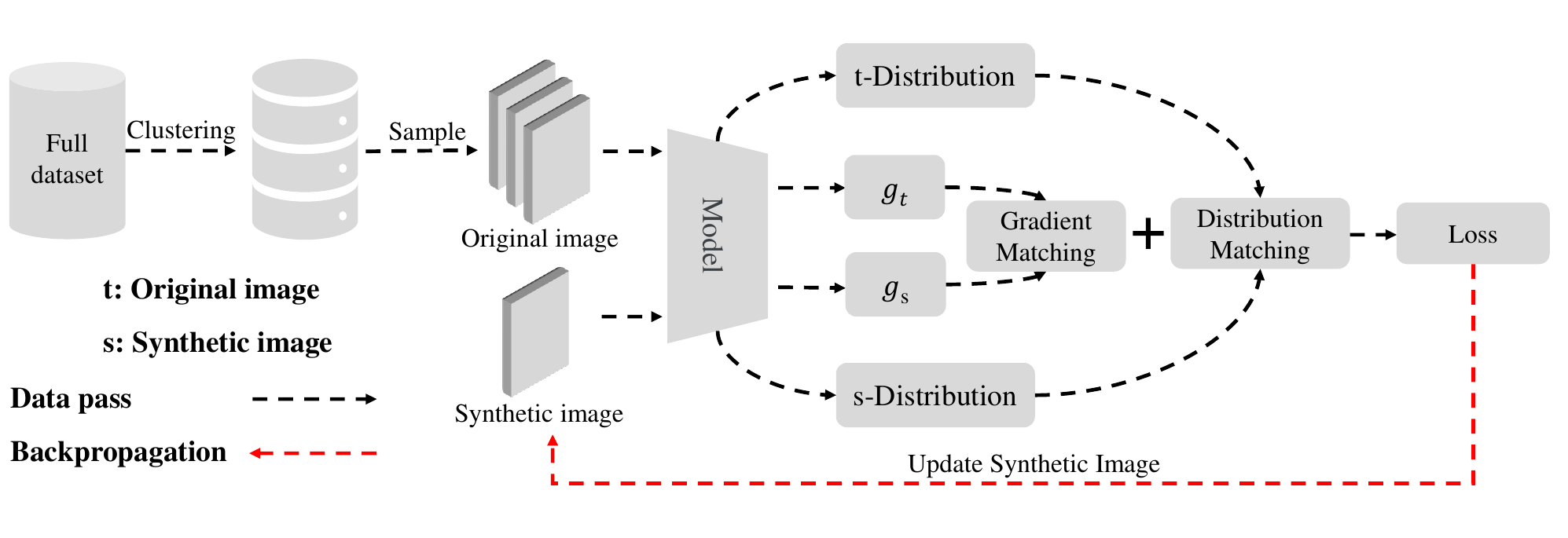}
\caption{The training pipeline of the proposed DREAM\raisebox{0.09\baselineskip}{+} strategy.}
\label{fig:pipeline}
\end{figure*}

\subsection{Preliminaries}
Given a large-scale dataset $\mathcal{T}=\{(\bm{x}_t^i,y_t^i)\}_{i=1}^{|\mathcal{T}|}$, the target of dataset distillation is to create a compact surrogate dataset $\mathcal{S}=\{(\bm{x}_s^i,y_s^i)\}_{i=1}^{|\mathcal{S}|}$ with minimal information loss, where $|\mathcal{S}|\ll |\mathcal{T}|$. Information loss is typically quantified by the performance drop observed when training a model on the original images $\mathcal{T}$ compared with the surrogate set $\mathcal{S}$.

Commonly adopted optimization-based methods follow a synthetic pipeline. Initially, the surrogate set $\mathcal{S}$ is initialized with randomly selected original images from $\mathcal{T}$. These synthetic images are then updated, guided by matching objectives $\phi(\cdot)$, to mimic the distribution of the original images. This process is formulated as follows:
\begin{equation}
\mathcal{S}^*=\arg\min_\mathcal{S}\mathbf{D}\left(\phi(\mathcal{S}),\phi(\mathcal{T})\right),
\end{equation}
where $\mathbf{D}$ stands for the matching metric. Typically, we opt for either training gradients or embedding features as the matching target, denoted as $\phi(\cdot)$. When we contemplate a random model $\mathcal{M}_\theta$ with training parameters $\theta$, the objective for $\mathcal{S}$ is to produce gradients that closely mirror those of $\mathcal{T}$ throughout the training process of $\mathcal{M}_\theta$ or embedding feature distributions identical to that of random $\mathcal{M}_\theta$. This objective can be expressed as:
\begin{subequations}
\label{eq:match}
\begin{equation}
    \mathcal{S}^*=\arg\min_\mathcal{S}\mathbf{D}\left(
    \nabla_\theta\mathcal{L}(\mathcal{M}_\theta(\mathcal{A}(\mathcal{S}))),
    \nabla_\theta\mathcal{L}(\mathcal{M}_\theta(\mathcal{A}(\mathcal{T})))\right),
\end{equation}
\centering \text{or}
\begin{equation}
    \mathcal{S}^*=\arg\min_\mathcal{S}\mathbf{D}\left(
    \xi(\mathcal{M}_\theta(\mathcal{A}(\mathcal{S}))),
    \xi(\mathcal{M}_\theta(\mathcal{A}(\mathcal{T})))\right),
\end{equation}
\end{subequations}
where $\mathcal{L}(\cdot,\cdot)$ represents the training loss, $\xi$ represents averaging the features in the channel dimension, and $\mathcal{A}$ denotes the differentiable augmentation techniques~\cite{karras2020training,zhao2020differentiable,tran2020towards,zhao2020image}.

In practice, the matching objectives are calculated on the synthetic images and a mini-batch of original images $\{(\bm{x}_t^i,y_t^i)\}_{i=1}^{N}$ sampled from $\mathcal{T}$ with the same class labels. The objective matching and training of $\mathcal{M}_\theta$ occur in an alternating manner. This process, involving the matching of gradients or embedding features at different training stages, constitutes the inner optimization loop of dataset distillation. The inner loop is iterated with different random $\mathcal{M}_\theta$ models to introduce diversity in matching gradients and features, denoted as the outer optimization loop.

Recent literature introduces a series of matching objectives that achieve significant test accuracy when trained on compact synthetic datasets~\cite{wang2022cafe,mtt,idc}. However, it is essential to note that the dataset distillation process itself remains time-consuming, indicating low training efficiency. In this work, we delve into the relationships between the training efficiency and the selection of original images utilized for matching, as well as the interplay between training efficiency and the chosen matching objectives. Drawing insights from the analysis, we introduce an innovative bidirectional matching strategy.

\subsection{Observations on Training Efficiency}

During the process of dataset distillation, knowledge is condensed by matching a subset of original images within a defined parameter space.
For memory restriction, there have to be sample selection for original images to form a mini-batch. 
The selection of these original images can significantly affect training efficiency, where recent literature usually adopts random sampling~\cite{zhao2020dc,idc}. 
Besides, the matching objectives can also influence the training efficiency.
Although there have been various objectives proposed, most of previous works rely on a single aspects among them~\cite{dm,idc}. 
Without losing the generality, here we use gradient matching as an example, and illustrate how these factors affect the efficient training of dataset distillation.

First, we examine the matching effect across samples from different distribution regions. 
Among all samples in a class, those closer to the center of the whole distribution tend to show higher prediction accuracy, indicating smaller backward gradients. 
In contrast, those located on the decision boundary show the opposite. 
In the case of gradient matching, center samples provide poor supervision contributions, while the gradients of boundary samples have a significant impact on the optimization direction. 
We show in Figure~\ref{fig:acc} the training accuracy curves for synthetic images matched only with center or boundary samples. 
It is obvious that the small gradients provided by the center samples quickly lose the guidance for the training process. 
Conversely, while boundary samples are crucial for delineating decision boundaries, relying solely on them for matching introduces chaotic matching targets, ultimately reducing the quality of the distillation process.

Second, we illustrate that random sampling does not guarantee a uniform distribution of samples inside mini-batches throughout the training process. We quantify this by recording the Maximum Mean Discrepancy (MMD) between the selected mini-batch and the overall class distribution during training, as shown in Figure~\ref{fig:mmd}. It can be observed that MMD remains at a consistently high level and has large fluctuations throughout the training process.

For gradient matching, when mini-batches cannot effectively and consistently cover the distribution of original samples, the gradient differences between individual samples become unbalanced. 
Due to the existence of boundary samples with large training gradients, the matching target of the mini-batch may be biased towards those samples, leading to unstable supervision. 
In addition, unevenly distributed small batches also mean that sample diversity is relatively limited. 
This imbalance is characterized by information redundancy in dense regions and scarcity of information in sparse regions, making mini-batches insufficient to represent the full width of the original data.

\begin{table*}[t]
\caption{Top-1 accuracy of test models trained on distilled synthetic images on multiple datasets. The distillation training is conducted with ConvNet-3. $^\dagger$ denotes the reported error range is reproduced by us. Best results are marked as \textcolor{red}{\textbf{red}}.  }
\label{tab:sota}
\centering
\small
\setlength{\tabcolsep}{7pt}
\begin{tabular}{c|ccc|ccc|ccc|ccc|ccc}
\toprule
\multirow{2}{*}{Dataset}&
\multicolumn{3}{c|}{MNIST} & \multicolumn{3}{c|}{FashionMNIST}  & \multicolumn{3}{c|}{SVHN} & \multicolumn{3}{c|}{CIFAR-10} & \multicolumn{3}{c}{CIFAR-100}\\
 & 1 & 10 & 50 & 1 & 10 & 50 & 1 & 10 & 50 & 1 & 10 & 50 & 1 & 10 & 50  \\ \midrule
Random & 64.9 & 95.1 & 97.9 & 51.4 & 73.8 & 82.5 & 14.6 & 35.1 & 70.9 & 14.4 & 26.0 & 43.4 & 4.2 & 14.6 & 30.0\\
Herding & 89.2 & 93.7 & 94.9 & 67.0 & 71.1 & 71.9 & 20.9 & 50.5 & 72.6 & 21.5 & 31.6 & 40.4 & 8.4 & 17.3 & 33.7\\
K-Center & 89.3 & 84.4 & 97.4 & 66.9 & 54.7 & 68.3 & 21.0 & 14.0 & 20.1 & 21.5 & 14.7 & 27.0 & - & - & -\\
Forgetting & 35.5 & 68.1 & 88.2 & 42.0 & 53.9 & 55.0 & 12.1 & 16.8 & 27.2 & 13.5 & 23.3 & 23.3 & 4.5 & 15.1 & 30.5\\
DD~\cite{wang2018dataset} & - & 79.5 & - & - & - & - & - & - & - & - & 36.8 & - & - & - & -\\
LD~\cite{LD} & 60.9 & 87.3 & 93.3 & - & - & - & - & - & - & 25.7 & 38.3 & 42.5 & 11.5 & - & - \\
DC~\cite{zhao2020dc} & 91.7 & 97.4 & 98.8 & 70.5 & 82.3 & 83.6 & 31.2 & 76.1 & 82.3 & 28.3 & 44.9 & 53.9 & 12.8 & 25.2 & - \\
DSA~\cite{zhao2021dsa} & 88.7 & 97.8 & 99.2 & 70.6 & 84.6 & 88.7 & 27.5 & 79.2 & 84.4 & 28.8 & 52.1 & 60.6 & 13.9 & 32.3 & 42.8 \\
DM~\cite{dm} & 89.7 & 97.5 & 98.6 & - & - & - & - & - & - & 26.0 & 48.9 & 63.0 & 11.4 & 29.7 & 43.6 \\
CAFE~\cite{wang2022cafe} & 93.1 & 97.2 & 98.6 & 77.1 & 83.0 & 84.8 & 42.6 & 75.9 & 81.3 & 30.3 & 46.3 & 55.5 & 12.9 & 27.8 & 37.9 \\
MTT~\cite{mtt} & - & - & - & - & - & - & - & - & - & 46.3 & 65.3 & 71.6 & 24.3 & 40.1 & 47.7 \\
IDC~\cite{idc} & 94.2 & 98.4 & 99.1 & 81.0 & 86.0 & 86.2 & 68.5 & 87.5 & 90.1 & 50.6 & 67.5 & 74.5 & - & 45.1 & - \\
IDM~\cite{zhao2023improved} & - & - & - & - & - & - & - & - & - & 45.6 & 58.6 & 67.5 & 20.1 & 45.1 & 50.0 \\
KIP~\cite{nguyen2020dataset,nguyen2021dataset} & 90.1 & 97.5 & 98.3 & 73.5 & 86.8 & 88.0 & 57.3 & 75.0 & 80.5 & 49.9 & 62.7 & 68.6 & 15.7 & 28.3 & -  \\ 
RFAD~\cite{loo2022efficient} & 94.4 & 98.5 & 98.8 & 78.6 & 87.0 & 88.8 & 52.2 & 74.9 & 80.9 & \textcolor{red}{\textbf{53.6}} & 66.3 & 71.1 & 26.3 & 33.0 & -  \\
HaBa~\cite{liu2022dataset} & 92.4 & 97.4 & 98.1 & - & - & - & 69.8 & 83.2 & 88.3 & 48.3 & 69.9 & 74.0 & \textcolor{red}{\textbf{33.4}} & 40.2 & 47.0  \\
FRePo~\cite{zhou2022dataset} & 93.0 & 98.6 & 99.2 & 75.6 & 86.2 & \textcolor{red}{\textbf{89.6}} & - & - & - & 46.8 & 65.5 & 71.7 & 28.7 & 42.5 & 44.3  \\
DREAM & 95.7 & 98.6 & 99.2 & 81.3 & 86.4 & 86.8 & 69.8 & 87.9 & 90.5 & 51.1 & 69.4 & 74.8 & 29.5 & 46.8 & 52.6 \\
DREAM\raisebox{0.14\baselineskip}{+} & \textcolor{red}{\textbf{96.1}} & \textcolor{red}{\textbf{98.6}} & \textcolor{red}{\textbf{99.2}} & \textcolor{red}{\textbf{82.6}} & \textcolor{red}{\textbf{87.2}} & 87.6 & \textcolor{red}{\textbf{71.8}} & \textcolor{red}{\textbf{88.9}} & \textcolor{red}{\textbf{91.5}} & 52.5 & \textcolor{red}{\textbf{69.9}} & \textcolor{red}{\textbf{75.3}} & 29.7 & \textcolor{red}{\textbf{47.4}} & \textcolor{red}{\textbf{52.6}} \\\bottomrule
\end{tabular}
\end{table*}

Furthermore, we claim that relying solely on a single optimization objective accesses limited information, which can also be improved for efficiency. 
Specifically, by only matching the training gradients, the consistency on feature distribution is overlooked. 
The feature-level matching typically produces more even distribution coverage over the original dataset~\cite{wang2022cafe}. 
With a more balanced synthetic data distribution, the gradient supervision can be better applied for optimization, and thereby further improves the training efficiency. 
In addition to the sample selection, the lack of feature alignment during the matching process also affects the efficiency of knowledge transfer.

These factors jointly lead to unstable optimization during the distillation process, ultimately reducing the training efficiency. We therefore advocate the development of a novel strategy to construct mini-batches with uniform and diverse distributions while optimizing the matching objective to achieve more efficient dataset distillation.

\subsection{Bidirectional Representative Matching}
For a stable and efficient optimization, we select representative original images for bidirectional matching. The selection process follows two basic principles. First, the selected images must be evenly distributed to prevent bias in the matching target. Second, while maintaining diversity, the selected samples should accurately reflect the overall sample distribution within the class.

To this end, we employ a clustering approach to select representative original images. Out of the considerations of uniform sub-cluster size and distribution, we use K-Means~\cite{k-means,k-means++,Omer_fast-pytorch-kmeans_2020} for sub-cluster partitioning. As shown in Figure~\ref{fig:cluster}, the clustering is performed within each class, generating $N$ sub-clusters that faithfully represent the sample density. Here, $N$ represents a predefined hyper-parameter of the mini-batch size of real images. The sub-cluster centers are strategically positioned to evenly distribute the entire class sample space,  and simultaneously hold sufficient diversity, perfectly meeting the above principles.

The entire training process is shown in Figure~\ref{fig:pipeline}. First, we randomly initialize a model and train it for one epoch. The initial training helps extract improved features for subsequent phases. The selected mini-batch of images as well as synthetic images with the same class labels are then passed through the model. This step produces embedding features and prediction scores. Next, we compute the classification loss and its corresponding gradient. In DREAM\raisebox{0.09\baselineskip}{+}, we adopt a distance metric $\mathbf{D}$ that combines embedding distance and gradient distance. The enhancement increases the efficiency of knowledge transfer throughout the process. The combined loss (as derived in Eq.~\ref{eq:match}) is back-propagated to update the synthetic image.

The sub-clusters are expected to have consistent information with the matching optimization. Therefore, we use the distillation model to extract the features for clustering as well as matching. At the same time, the model is updated in the inner loop  to provide more diverse gradient supervision for matching at each stage. To account for the additional time cost, the clustering process is performed every $I_{int}$ iterations.

Furthermore, similar to~\cite{cui2022dc}, we apply a clustering at the beginning of the training process. Here, we cluster the data in each class into subclusters, each subcluster corresponding to a predefined number of images per class. The center sample of each sub-cluster is selected as the initialization point of the synthesized image. This balanced, cluster-based initialization method better captures data distribution and accelerates convergence from the beginning of the training process.

\section{Experiments}

\begin{figure*}[t]
\centering
\begin{subfigure}{0.32\textwidth}
    \includegraphics[width=\textwidth]{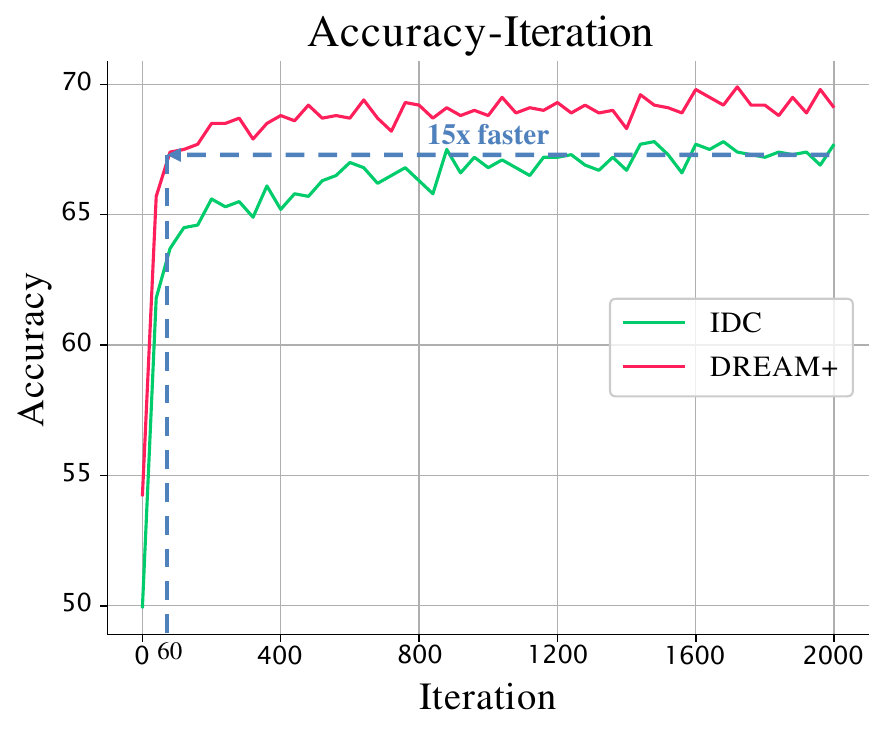}
    \caption{The accuracy curve of adding DREAM\raisebox{0.09\baselineskip}{+} to IDC(gradient). }
\end{subfigure}
\hfill
\begin{subfigure}{0.32\textwidth}
    \includegraphics[width=\textwidth]{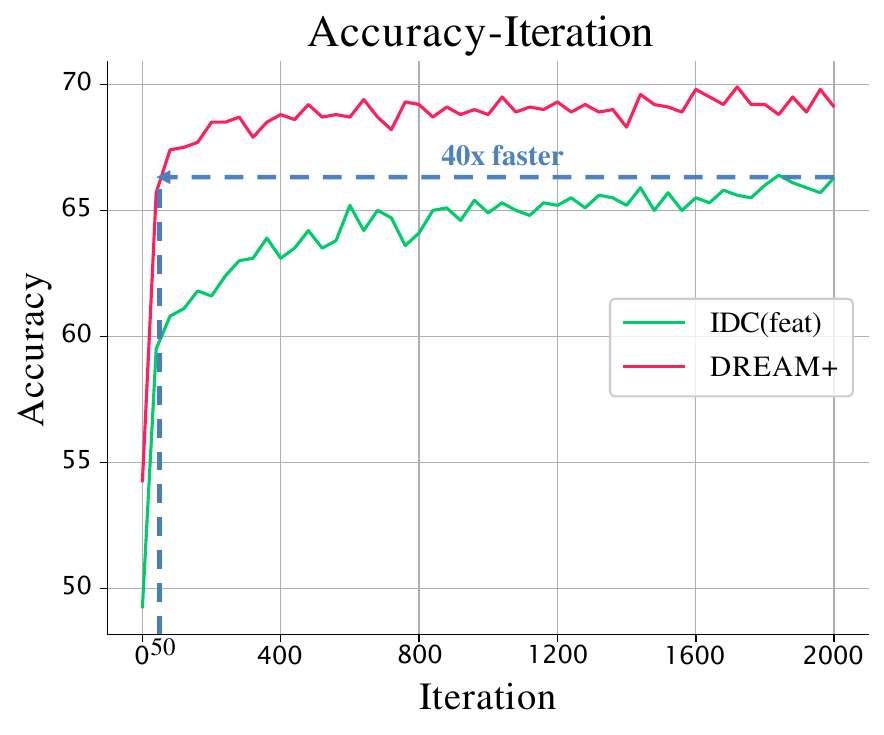}
    \caption{The accuracy curve of adding DREAM\raisebox{0.09\baselineskip}{+} to IDC(feature). }
    \label{fig:curve-dm}
\end{subfigure}
\hfill
\begin{subfigure}{0.32\textwidth}
    \includegraphics[width=\textwidth]{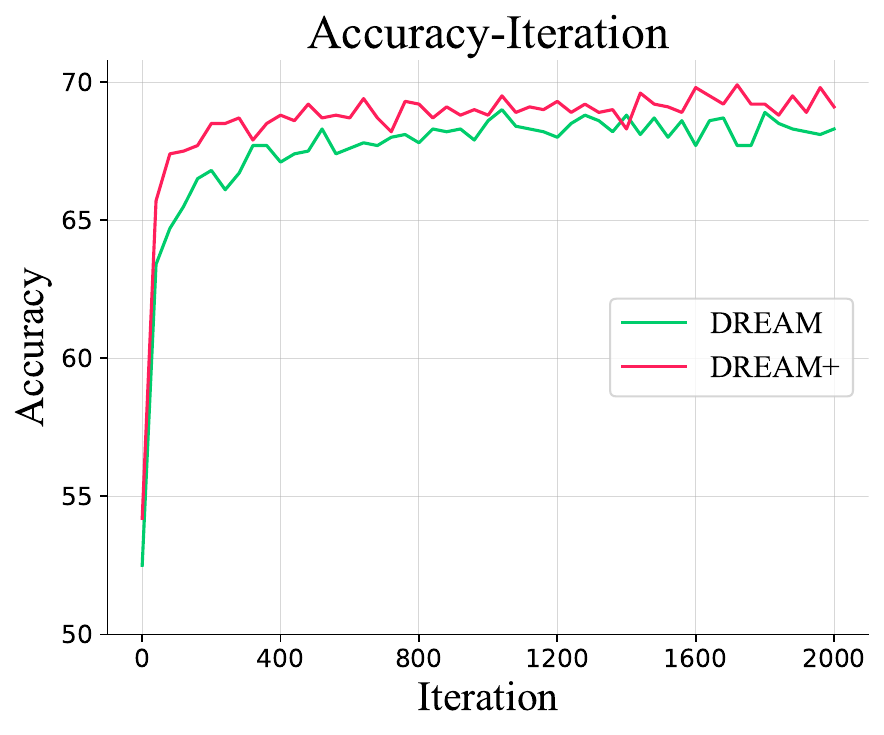}
    \caption{The accuracy curves of DREAM\raisebox{0.09\baselineskip}{+} and DREAM. }
\end{subfigure}
\caption{Applying the DREAM\raisebox{0.09\baselineskip}{+} strategy brings stable performance and efficiency improvements. }
\label{fig:curve1}
\end{figure*}

\subsection{Datasets and Implementation Details}

We validate the effectiveness of our method on several popular datasets, including CIFAR-10~\cite{krizhevsky2009learning}, CIFAR-100~\cite{krizhevsky2009learning}, SVHN~\cite{netzer2011reading}, MNIST~\cite{lecun1998gradient}, FashionMNIST~\cite{xiao2017fashion}, and TinyImageNet~\cite{deng2009imagenet}. Our evaluation involves training a model on the distilled synthetic images and testing it on the original testing images. We report Top-1 accuracy to demonstrate performance.

Unless otherwise specified, we employ 3-layer convolutional networks (ConvNet-3)~\cite{gidaris2018dynamic} with 128 filters and instance normalization~\cite{ulyanov2016instance}. The matching mini-batch size for original images is set to 128. In the case of TinyImageNet, where the image resolution is 64$\times$64, we utilize ConvNet-4. Our default baseline method is IDC~\cite{idc}. The matching objective combines gradient matching and distribution matching. The matching metric $\mathbf{D}$ in Eq.~\ref{eq:match} is empirically defined as the mean squared error for CIFAR-10, CIFAR-100, TinyImageNet, and SVHN. For MNIST and FashionMNIST, we set $\mathbf{D}$ as the mean absolute error~\cite{idc}. We perform a total of 1,200 matching iterations, with each iteration comprising 100 inner loops. We employ SGD as the optimizer, with a learning rate set to 0.005.

For clustering, we employ the distillation model for feature extraction. 
The clustering interval $I_{int}$ is set as 10 iterations, whose sensitiveness is analyzed in Sec.~\ref{ana:interval}. 
We also analyze the influence of different sampling strategy in Sec.~\ref{ana:sampling}. 
For evaluation, we train a network for 1,000 epochs on the distilled images with a learning rate of 0.01. 
We conduct 5 runs for each experiment and report the mean and standard deviation of the results.

\begin{table}[t]
\caption{Top-1 accuracy of test models trained on distilled synthetic images on TinyImageNet. The distillation training is conducted with ConvNet-4. }
\label{tab:tiny}
\centering
\small
\setlength{\tabcolsep}{4pt}
\begin{tabular}{cc|ccc|c}
\toprule
IPC & Ratio \% & DM~\cite{dm} & MTT~\cite{mtt} & \textbf{DREAM\raisebox{0.10\baselineskip}{+}}& Whole \\ \midrule
1 & 0.017 & 3.9$_{\pm0.2}$ & 8.8$_{\pm0.3}$ & \bf{10.5}$_{\pm0.4}$ & \\
10 & 0.17 & 12.9$_{\pm0.4}$ & 23.2$_{\pm0.2}$ & \textbf{24.0}$_{\pm0.4}$ & 37.6$_{\pm0.4}$ \\
50 & 0.83 & 24.1$_{\pm0.3}$ & 28.0$_{\pm0.3}$ & \bf{29.5}$_{\pm0.3}$ &  \\ \bottomrule
\end{tabular}
\end{table}

\subsection{Comparison with State-of-the-art Methods}

We perform a comprehensive comparison of DREAM\raisebox{0.09\baselineskip}{+} with state-of-the-art (SOTA) coreset-based and optimization-based methods across multiple datasets, each with varying images-per-class (IPC) settings, as summarized in Tab.~\ref{tab:sota}. Additionally, for the TinyImageNet dataset, we specifically compare DREAM\raisebox{0.09\baselineskip}{+} with DM~\cite{dm} and MTT~\cite{mtt} as presented in Tab.~\ref{tab:tiny}. 
DREAM\raisebox{0.09\baselineskip}{+} consistently demonstrates state-of-the-art (SOTA) results across most cases. 
The reduced performance gap between the small-scale distilled dataset and its original dataset means less information loss during the dataset distillation process.
It illustrates the effectiveness of bidirectional matching of representative samples in our method.
It is worth mentioning that RFAD~\cite{loo2022efficient} employs a ConvNet with 1024 convolutional channels, while the results we report are based on a 128-channel ConvNet. DREAM\raisebox{0.09\baselineskip}{+} outperforms RFAD in extracting better synthetic images except for IPC=1 on CIFAR-10.
Meanwhile, HaBa~\cite{liu2022dataset} incorporates a data hallucination process that generates additional samples from the base image. HaBa achieves superior performance with 1 IPC on CIFAR-100. However, in other cases, DREAM\raisebox{0.09\baselineskip}{+} consistently gains superior performance compared to HaBa.
These results together highlight the competitive performance of DREAM\raisebox{0.09\baselineskip}{+} under different IPC settings across multiple datasets.

\begin{table}[t]
\setlength{\tabcolsep}{2.5pt}
    \caption{Ablation study on the components of the proposed DREAM. RM indicates Representative Matching, Init stands for clustering-based initialization, and BM indicates Bidirectional Matching. “Iter” stands for the required iterations to achieve the baseline performance.}
    \label{tab:component}
    \centering
    \small
    \begin{tabular}{lccccc|lcccc}
    \toprule
        \multirow{2}{*}{}& \multicolumn{3}{c}{Comp} & \multirow{2}{*}{Top-1} & \multirow{2}{*}{Iter} & \multirow{2}{*}{}& \multicolumn{3}{c}{Comp} & \multirow{2}{*}{Top-1} \\
         & RM & Init & BM & & & & RM & Init & BM\\
        \midrule
         \multirow{6}{*}{IDC}& - & - & -&67.5$_{\pm0.5}$ & 1000 & \multirow{3}{*}{DC} & - & - & - &44.9$_{\pm0.5}$\\
          & \checkmark & - & - &68.9$_{\pm0.5}$ & 350 & & \checkmark & \checkmark &  -&45.9$_{\pm0.3}$ \\
          & - & \checkmark & - &68.1$_{\pm0.3}$ & 750 & & \checkmark &\checkmark &\checkmark & \textbf{46.9}$_{\pm0.4}$\\\cline{7-11}
          & - & - & \checkmark & 68.7$_{\pm0.3}$ & 480 & \multirow{3}{*}{DSA} & - & - & - & 52.1$_{\pm0.5}$\\
          & \checkmark & \checkmark & - & 69.4$_{\pm0.4}$ & 150 &  & \checkmark & \checkmark &  -& 53.1$_{\pm0.4}$\\
          & \checkmark & \checkmark &  \checkmark&\textbf{69.9}$_{\pm0.5}$ & \textbf{60} & & \checkmark & \checkmark & \checkmark &\textbf{53.5}$_{\pm0.2}$\\
        \bottomrule
    \end{tabular}
\end{table}
\subsection{Efficiency comparison}
We evaluate the efficiency of our proposed method on both gradient-based and feature-based IDC, as shown in Figure~\ref{fig:curve1}. Notably, our approach significantly reduces the number of iterations required for dataset distillation. Among them, for gradient matching, DREAM\raisebox{0.09\baselineskip}{+} reduces the number of iterations by more than 15 times; for distribution matching, DREAM\raisebox{0.09\baselineskip}{+} reduces the number of iterations by more than 40 times. As the training iterations increases, DREAM\raisebox{0.09\baselineskip}{+} further boost the performance of the model. 
Additionally, the improved version of DREAM\raisebox{0.09\baselineskip}{+} also demonstrates better efficiency compared with previous DREAM. 
This empirical evidence highlights the effectiveness of our bidirectional representative matching in improving the stability and efficiency of dataset distillation.

\subsection{Ablation Study and Analysis}
We conduct comprehensive experiments to evaluate the effectiveness of our proposed DREAM\raisebox{0.09\baselineskip}{+} strategy. By default, the experiments are conducted at 10 IPC settings on CIFAR-10.

\textbf{Component Combination Evaluation. }
We first perform an analysis on the components of the proposed DREAM\raisebox{0.09\baselineskip}{+} strategy in Table~\ref{tab:component}.
Representative matching and bidirectional matching greatly reduce the number of iterations required to reach baseline performance.
Furthermore, clustering-based initialization shows a huge performance advantage before training starts, although its final impact is still relatively limited. However, when combined with bidirectional representational matching, it provides stable enhancement and accelerates the training convergence.
By integrating all these components, the full DREAM\raisebox{0.09\baselineskip}{+} approach proved highly effective, reducing the number of iterations required to achieve baseline performance by more than 15 times.
These findings highlight the importance of representative matching and bidirectional matching components in improving the training efficiency and dataset distillation performance.

\begin{table}[t]
    \caption{Ablation study on cross architecture distilled dataset performance of the proposed DREAM strategy. The dataset is first distilled on a model $\mathrm{D}$ and then validated on another model $\mathrm{T}$. $^\dagger$ denotes the accuracy is reproduced by us. }
    \label{tab:crossarch}
    \centering
    \small
    \begin{tabular}{lcccc}
    \toprule
        & $\mathrm{D}\backslash\mathrm{T}$ & Conv-3 & Res-10 & Dense-121 \\
        \midrule
        \multirow{2}{*}{MTT~\cite{mtt}} & Conv-3 & 64.3$_{\pm0.7}$ & 34.5$_{\pm0.6}$$^\dagger$ & 41.5$_{\pm0.5}$$^\dagger$\\
         & Res-10 & 44.2$_{\pm0.3}$$^\dagger$& 20.4$_{\pm0.9}$$^\dagger$ & 24.2$_{\pm1.3}$$^\dagger$\\
        \midrule
        \multirow{2}{*}{IDC~\cite{idc}} & Conv-3 & 67.5$_{\pm0.5}$ & 63.5$_{\pm0.1}$ & 61.6$_{\pm0.6}$ \\
         & Res-10 & 53.6$_{\pm0.6}$$^\dagger$ & 50.6$_{\pm0.9}$$^\dagger$ & 51.7$_{\pm0.6}$$^\dagger$\\
         \midrule
        \multirow{2}{*}{DREAM~\cite{liu2023dream}} & Conv-3 & 69.4$_{\pm0.5}$ & 66.3$_{\pm0.8}$ & 65.9$_{\pm0.5}$ \\
         & Res-10 & 53.7$_{\pm0.6}$$^\dagger$ & 51.0$_{\pm0.9}$$^\dagger$ & 52.8$_{\pm0.6}$$^\dagger$\\
        \midrule
        \multirow{2}{*}{DREAM\raisebox{0.09\baselineskip}{+}} & Conv-3 & \textbf{69.9}$_{\pm0.5}$ & \textbf{66.5}$_{\pm0.8}$ & \textbf{66.0}$_{\pm0.5}$\\
         & Res-10 & \textbf{53.8}$_{\pm0.6}$ & \textbf{51.2}$_{\pm0.9}$ & \textbf{53.0}$_{\pm0.6}$\\
        \bottomrule
    \end{tabular}
\end{table}

\begin{table}[t]
    \caption{Ablation study on different sampling strategy to form a mini-batch from sub-clusters. }
    \centering
    \small
    \setlength{\tabcolsep}{4pt}
    \begin{tabular}{lc|cccc}
    \toprule
        \multirow{2}{*}{DREAM} & \multirow{2}{*}{} & \multicolumn{4}{c}{Sub-cluster number $N$} \\
        & & 32 & 64 & 128 & 256 \\
        \midrule
         & 1 & 67.2$_{\pm0.3}$ & 68.5$_{\pm0.1}$ & \textbf{69.4}$_{\pm0.4}$ & 68.9$_{\pm0.2}$ \\
        Samples per& 2 & 67.7$_{\pm0.3}$ & 68.6$_{\pm0.3}$ & 69.2$_{\pm0.7}$ & - \\
        sub-cluster $n$ & 4 & 67.7$_{\pm0.4}$ & 68.7$_{\pm0.4}$ & - & - \\
        & 8 & 67.5$_{\pm0.3}$ & - & - & - \\
    \bottomrule
    \end{tabular}
    
    \vspace{10pt}
    
    \begin{tabular}{lc|cccc}
    \toprule
        \multirow{2}{*}{DREAM+} & \multirow{2}{*}{} & \multicolumn{4}{c}{Sub-cluster number $N$} \\
        & & 32 & 64 & 128 & 256 \\
        \midrule
         & 1 & 67.4$_{\pm0.3}$ & 69.4$_{\pm0.1}$ & \textbf{69.9}$_{\pm0.4}$ & 69.6$_{\pm0.2}$ \\
        Samples per& 2 & 68.7$_{\pm0.3}$ & 69.8$_{\pm0.3}$ & 69.8$_{\pm0.7}$ & - \\
        sub-cluster $n$ & 4 & 68.8$_{\pm0.4}$ & 69.6$_{\pm0.4}$ & - & - \\
        & 8 & 69.0$_{\pm0.3}$ & - & - & - \\
    \bottomrule
    \end{tabular}
    \label{tab:sampling}
\end{table}
To further emphasize the effectiveness of bidirectional representative matching, we show the results in Figure~\ref{fig:method}. The figure visually demonstrates how our strategy affects the training efficiency and synthetic dataset performance.
As shown in Figure~\ref{fig:acc}, we obtain significant improvements in both performance and efficiency by simply using samples sampled from sub-clusters as matching targets (previous DREAM strategy). 
The bidirectional representative matching further enhances the acceleration, achieving baseline performance in less than one-fifteenth of the original required number of training iterations.
Furthermore, by increasing the number of training iterations, the mutually constrained matching objectives of gradient and feature matching enhance the representation capabilities of synthetic data, leading to an overall improvement in the performance.

For the sample distribution, as shown in Figure~\ref{fig:mmd}, we calculate the MMD to the real data distribution of images selected by our method and random sampling. 
The results consistently show that the former has lower MMD scores and less fluctuations. 
The reduction in fluctuations indicates that the sub-cluster centers effectively and stably cover the feature distribution, and thereby reduce the sample-level noise during training. 
With sufficient sample diversity, uniform distribution, and appropriate bidirectional supervision, DREAM\raisebox{0.09\baselineskip}{+} ensures that the optimization process of dataset distillation training is smoother and more robust.
To further illustrate the generality of DREAM\raisebox{0.09\baselineskip}{+}, we apply representative bidirectional matching and clustering-based initialization to several other baseline methods. 
The results, shown in Tab.~\ref{tab:component}, demonstrate similar improvements. 
It confirms that DREAM\raisebox{0.09\baselineskip}{+} is suitable for a variety of dataset distillation frameworks and can significantly improve the training efficiency.

\begin{figure}[t]
    \centering
    \includegraphics[width=0.45\textwidth]{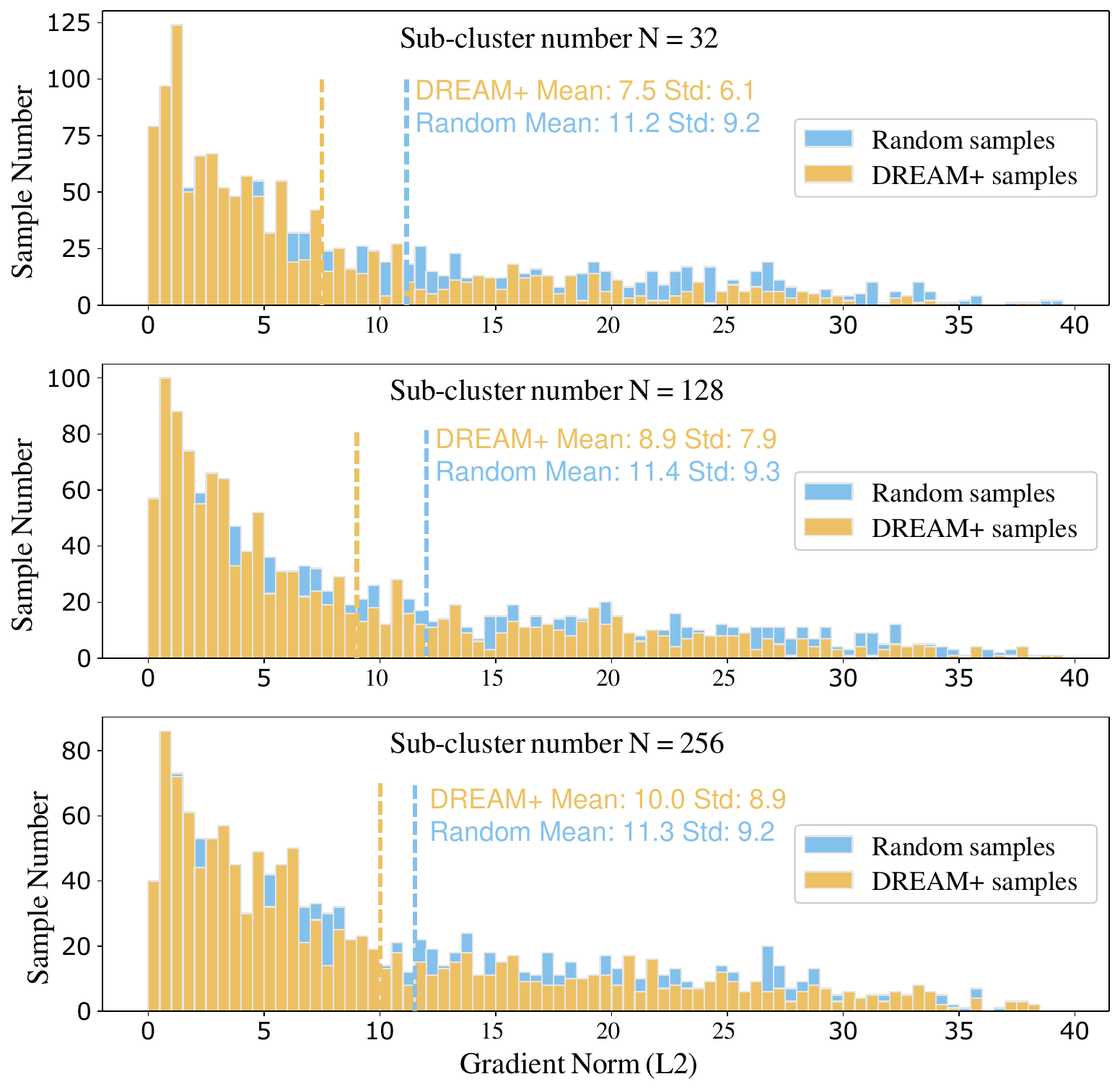}
    \caption{The gradient distribution comparison between random sampling and our proposed DREAM strategy under different sub-cluster sample number $N$. Best viewed in color.  }
    \label{fig:sampling}
\end{figure}

\begin{figure*}
\centering
\begin{subfigure}{0.24\textwidth}
    \includegraphics[width=\textwidth]{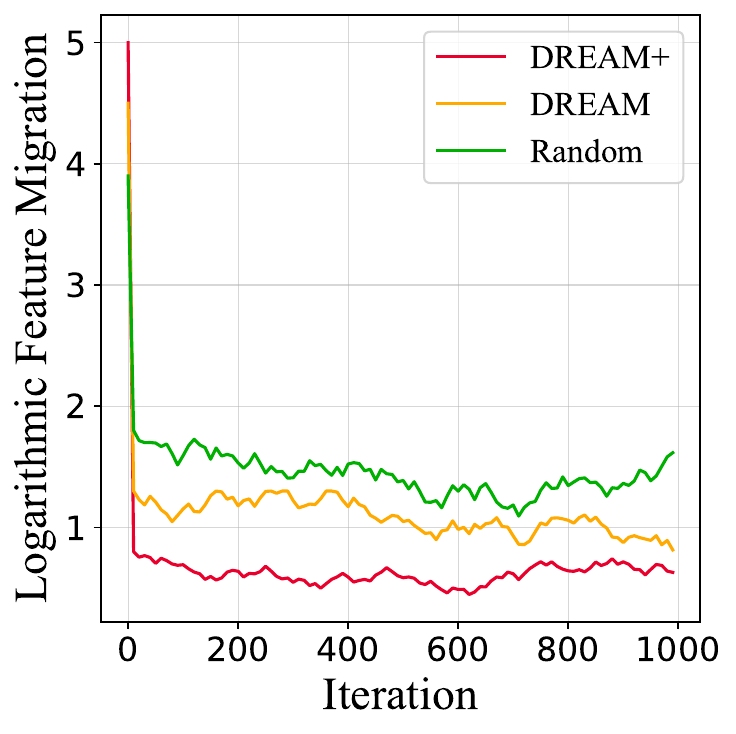}
    \caption{}
    \label{fig:osci}
\end{subfigure}
\begin{subfigure}{0.24\textwidth}
    \includegraphics[width=\textwidth]{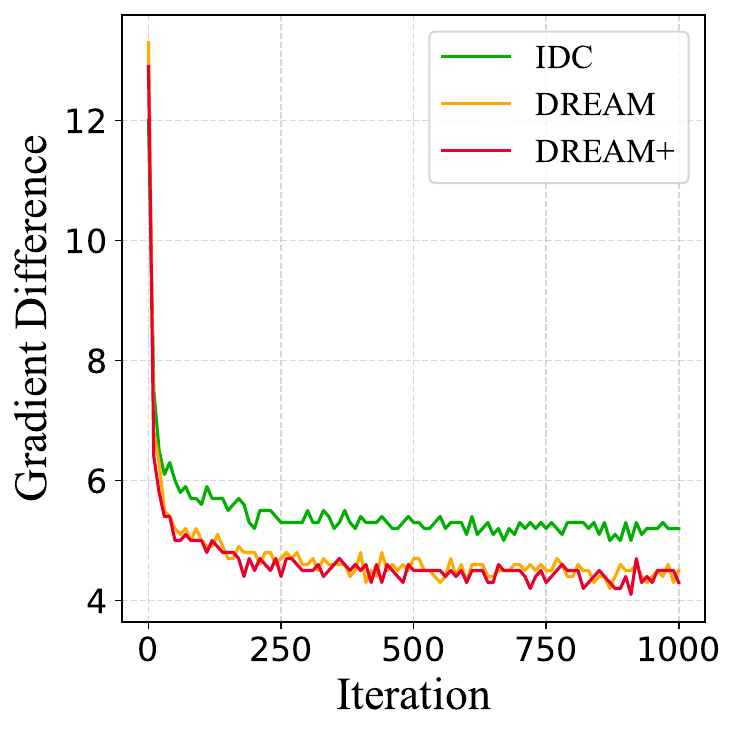}
    \caption{}
    \label{fig:loss-curve}
\end{subfigure}
\begin{subfigure}{0.24\textwidth}
    \includegraphics[width=\textwidth]{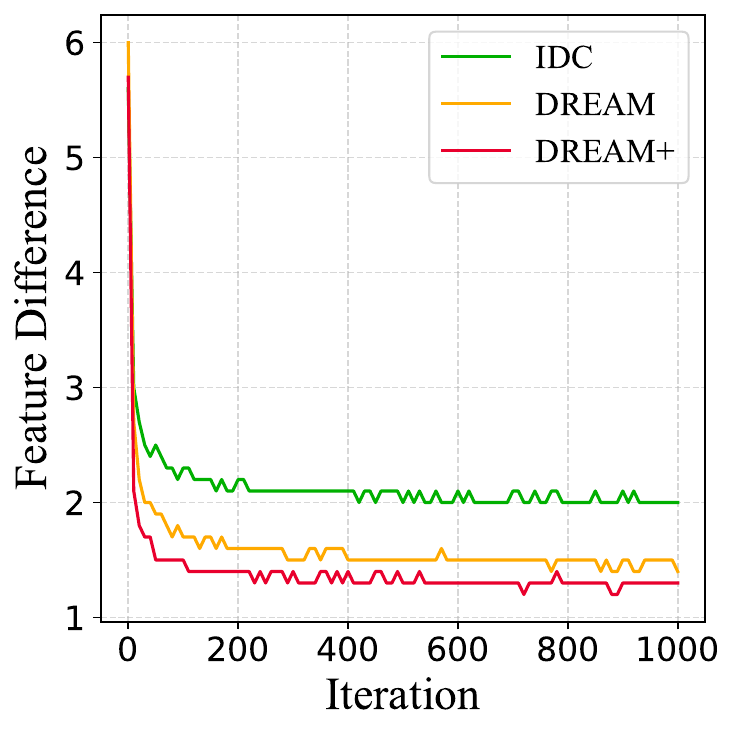}
    \caption{}
    \label{fig:loss-curve2}
\end{subfigure}
\begin{subfigure}{0.24\textwidth}
    \includegraphics[width=\textwidth]{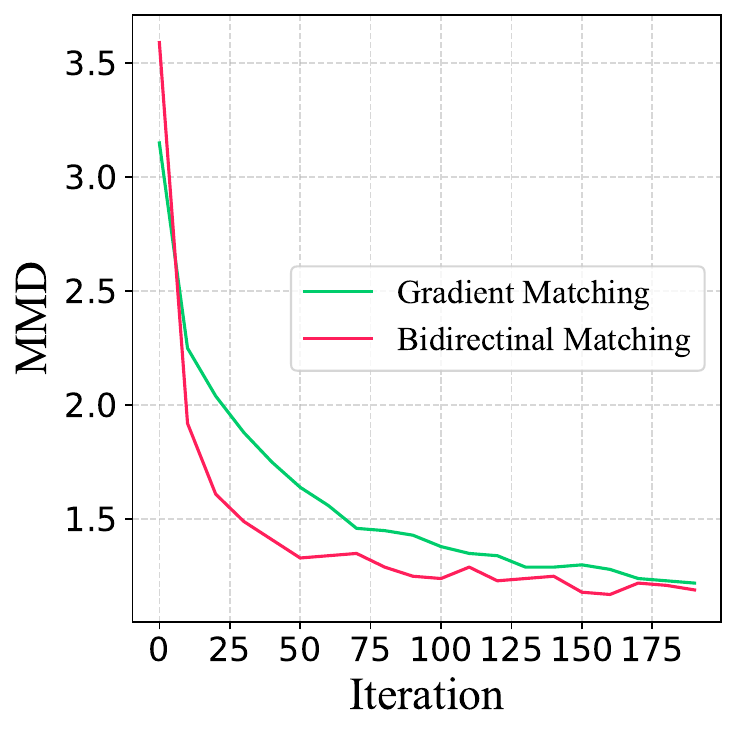}
    \caption{}
    \label{fig:feature_mmd}
\end{subfigure}

\caption{(a): The feature migration during the training process. (b): Curve of MMD variation between synthetic and original images. (c): The gradient difference curve during the training process. (d): The feature difference curve during the training process.  }
\end{figure*}

\textbf{DREAM\raisebox{0.09\baselineskip}{+} on Distribution Matching. }
In addition to gradient matching, we also explore the adaptability of our method to feature distribution matching. Random sampling not only introduces biased matching targets in gradient matching, it also has a similar impact on distribution matching. 
Specifically, random sampling tends to select samples around the center of feature distribution. It would reduce the feature diversity and training efficiency.

We conduct the experiments based on IDC~\cite{idc} under the setting of 10 images of each class on the CIFAR-10. 
The original IDC method performs significantly worse than gradient matching, which is consistent with the conclusion drawn by the previous work~\cite{idc}. 
DREAM\raisebox{0.09\baselineskip}{+} substantially improves in the performance and requires only about one-fortieth of iteration number to reach the baseline performance, as shown in Figure~\ref{fig:curve-dm}.

\textbf{Cross Architecture Generalization Analysis. }
A recurring challenge in dataset distillation methods is their inability to generalize effectively across different architectures. 
This limitation is due to the fact that synthetic images tend to be over-fitted to the specific architecture used for matching~\cite{zhao2020dc,idc}.
To evaluate the cross-architecture performance of our proposed DREAM\raisebox{0.09\baselineskip}{+} strategy, we conducted the experiments in Table~\ref{tab:crossarch}. 
The compact datasets are distilled with ConvNet-3 and ResNet-10~\cite{he2016deep}, and then evaluated on ConvNet-3, ResNet-10, and DenseNet-121~\cite{huang2017densely}.

It is worth noting that DREAM\raisebox{0.09\baselineskip}{+} not only outperforms other methods on specific distillation models, but also achieves significant performance improvements on other unseen network architectures. This strong cross-architecture generalization highlights that DREAM\raisebox{0.09\baselineskip}{+} builds representations of datasets with clearer decision boundaries than random sampling. Synthetic data that has closer overall distribution to the original data helps the model learn more general features and knowledge.

\textbf{Sampling Strategy Analysis. }
\label{ana:sampling}
We delve into the impact of different sampling strategies on the training results, as performed in Table~\ref{tab:sampling} and Figure~\ref{fig:sampling}. Our representative matching approach entails clustering for each class and subsequently sampling original images from sub-clusters to form mini-batches.
By varying the sub-cluster number and selected sample number per sub-cluster, we generate original image mini-batches that differ in scale and diversity. 
The ablation study provides better interpretability for the effectiveness of our approach.

In general, the performance of the dataset significantly benefits from representative matching compared to the baseline (67.5). However, specific nuances become apparent upon closer examination.
For instance, with a small sub-cluster number (e.g., $N=32$), the sub-cluster centers tend to be concentrated in regions with smaller gradients, as depicted in the first row of Figure~\ref{fig:sampling}. 
As the random model $\mathcal{M}_\theta$ undergoes training, these samples gradually lose their ability to provide effective gradient-based supervision, ultimately resulting in sub-optimal performance.
Conversely, a larger sub-cluster number (e.g., $N=256$) leads to a distribution that closely resembles random sampling, and causes a minor performance drop. 
Due to memory constraints, further increasing $N$ is unfeasible, but it is reasonable to assume that extreme conditions would yield results similar to those of random sampling.

On the other hand, variations in the sample number per sub-cluster ($n$) appear to exert only a marginal influence on results. 
The configuration involving one center sample per sub-cluster and a total of 128 sub-clusters provides optimal gradient-based supervision, as evidenced by the second row of Figure~\ref{fig:sampling}. 
Consequently, this configuration is selected for mini-batch composition.
In addition, under different sampling strategies, the DREAM\raisebox{0.09\baselineskip}{+} strategy has improved to a certain extent compared with DREAM, which also verifies the effectiveness of the new method.

\textbf{Training Stability Analysis. }
In order to describe the impact of DREAM\raisebox{0.09\baselineskip}{+} on the training process more intuitively, we visualize the feature migration during the distillation process of DREAM\raisebox{0.09\baselineskip}{+}, DREAM and random sampling.
We save the distilled images at intervals of 20 iterations and use a well-trained network to extract the features. 
The Euclidean distance between consecutive versions of the image is calculated and summarized in Figure~\ref{fig:osci}.

For both DREAM\raisebox{0.09\baselineskip}{+} and DREAM, the synthetic images go through a large feature migration in the early stages of training, which fully demonstrates that representative matching accelerates the optimization process of synthetic images. 
When the number of iterations is slightly increased, the migration of methods based on representative matching already turns small and stable.
It illustrates that representative sampling efficiently optimizes the images to a relatively optimal position, and makes subsequent fine-tuning. 
For synthetic images that match randomly sampled original images, the feature migration remains relatively high. 
This phenomenon is partly attributed to the uneven mini-batch of noisy matching targets, where the optimization is biased by the large-gradient samples inside a mini-batch, hindering a stable optimization process.
Besides, compared with DREAM, DREAM\raisebox{0.09\baselineskip}{+} shows further improvement in providing a stable overall feature migration, which indicates that distribution matching also effectively constrains the optimization process.

\begin{figure}
\begin{subfigure}{0.24\textwidth}
    \includegraphics[width=\textwidth]{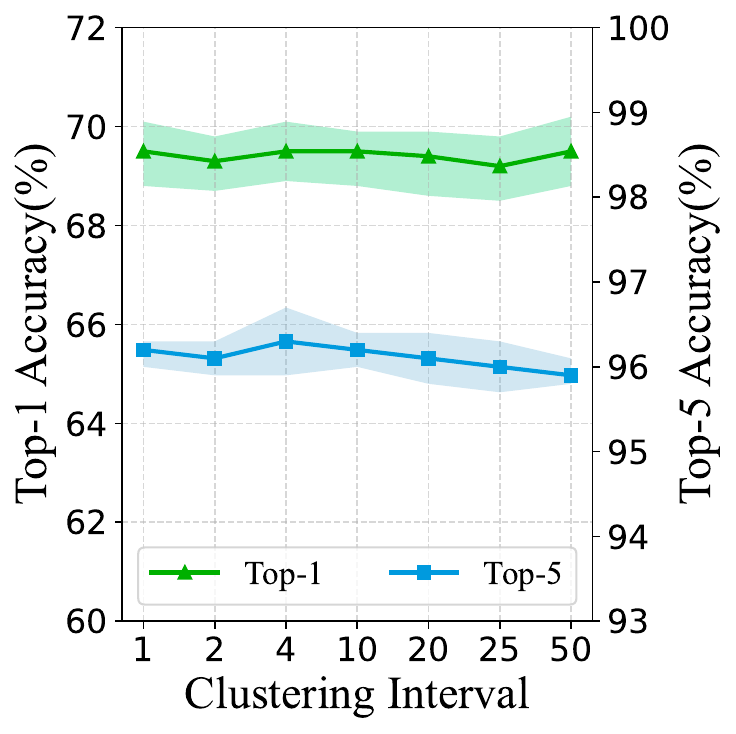}
    \caption{}
    \label{fig:interval}
\end{subfigure}
\begin{subfigure}{0.24\textwidth}
    \includegraphics[width=\textwidth]{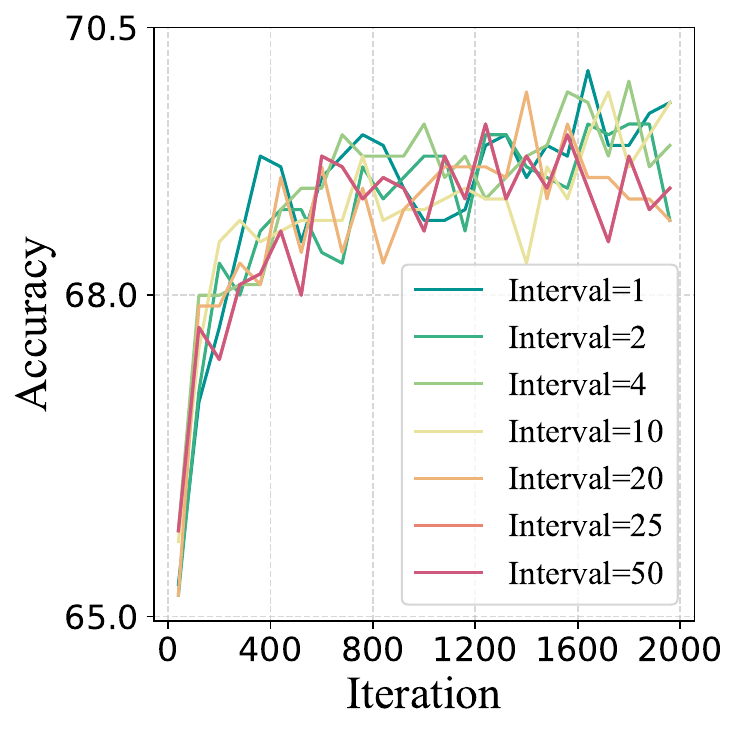}
    \caption{}
    \label{fig:interval_accuracy}
\end{subfigure}
\caption{Ablation study on different clustering interval. As the clustering interval increases, the efficiency of distillation gradually decreases.}
\end{figure}

\textbf{Clustering Interval Sensitivity Analysis. }
\label{ana:interval}
We further analyze the sensitivity of the clustering interval $I_{int}$. 
Our findings are shown in Figure~\ref{fig:interval}. 
We observe that in representative bidirectional matching, different clustering intervals have little impact on top-1 accuracy and top-5 accuracy.
We also visualize the iteration-accuracy curve under different clustering interval settings in ~\ref{fig:interval_accuracy}. It can be found that when the clustering interval gradually increases, the efficiency of distillation gradually decreases, but overall there is no significant impact on performance.
Based on these observations, we choose a clustering interval of 10. 
This choice achieves a balance between performance on synthetic datasets and the additional computational time introduced by clustering.

\begin{table}[t]
    \caption{Time cost of adding DREAM strategy (s).}
    \label{tab:time}
    \centering
    \small
    \begin{tabular}{lcccc}
    \toprule
        \multirow{2}{*}{Datasets} & \multirow{2}{*}{Methods} & \multirow{2}{*}{Clustering} & Update & Inner \\
        & & & Images & Loop \\
        \midrule
        \multirow{2}{*}{CIFAR-10}& IDC~\cite{idc} & - & 0.2 & 0.2 \\ 
        & DREAM+ & 0.1 & 0.2 & 0.3 \\
        \midrule
        \multirow{2}{*}{CIFAR-100}& IDC~\cite{idc} & - & 2.0 & 2.0 \\ 
        & DREAM+ & 0.1 & 2.0 & 2.1 \\
        \bottomrule
    \end{tabular}
\end{table}
\textbf{Clustering Analysis. }
To provide a comprehensive perspective on the computational impact of the clustering process, we present the extra time costs incurred in Table~\ref{tab:time}.
For CIFAR-10, each inner loop involves both the matching process and image updating, which collectively consume approximately 0.2 seconds. 
Every ten inner loops, we introduce a clustering process, which requires an additional 1 second. 
By average, this translates to a clustering time of 0.1 seconds per inner loop. 
Consequently, the total average duration of an inner loop becomes 0.3 seconds, compared to the original 0.2 seconds.
Compared with DREAM, since the features used in the newly introduced distribution alignment of DREAM\raisebox{0.09\baselineskip}{+} come from the features that have been calculated in the forward pass in gradient matching, the newly introduced time overhead is very small and can be ignored.
Remarkably, considering that we achieve the same level of performance with only one-twentieth to one-tenth of the iterations, this implementation of DREAM\raisebox{0.09\baselineskip}{+} allows us to save over 85\% of the time. For CIFAR-100, which involves a more extensive set of classes, the extra clustering time accounts for a mere twentieth of the original image updating time, which is negligible.
In essence, DREAM\raisebox{0.09\baselineskip}{+} contributes significantly to enhanced the training efficiency, and substantially reducing the required training time for dataset distillation.

\subsection{Visualizations}

\textbf{Gradient Difference Curve.}
Given that dataset distillation training depends on the gradient matching to some extent, and the smaller the gradient difference indicates the more effective the matching, we show the gradient difference curve of the dataset distillation process in Figure~\ref{fig:loss-curve}. 
The gradient difference is calculated based on the training loss, as defined in the Eq.~\ref{eq:match}.
We compared the DREAM\raisebox{0.09\baselineskip}{+} curve with IDC and DREAM. Across the entire training trajectory, DREAM\raisebox{0.09\baselineskip}{+} exhibits smaller gradient differences compared to baseline methods. This observation serves a dual purpose.
First, it confirms the efficacy of DREAM\raisebox{0.09\baselineskip}{+} in improving training efficiency, successfully reducing the gradient difference within a limited number of iterations. Second, the large fluctuations seen in the baseline method confirm the existence of noise gradients generated by random sampling.

\textbf{Feature Difference Curve.}
Feature distribution matching is another critical aspect in the training process of dataset distillation. The smaller the feature distribution difference means that the synthetic data more accurately approximates the feature distribution of the original data. This in turn helps the model acquire more general features and knowledge.
In Figure~\ref{fig:loss-curve2} and Figure~\ref{fig:feature_mmd} we provide a visualization of the feature difference during the training and the MMD between the feature distributions of the synthetic and original data throughout the distillation process.
Compared with pure gradient matching, the introduction of bidirectional matching leads to better feature alignment, especially in the early stages of training. The bidirectional optimization strategy effectively enhances the stability and efficiency of dataset distillation and alleviates potential problems related to feature transformation.

\begin{figure}
    \centering
    \begin{overpic}[width=0.48\textwidth]{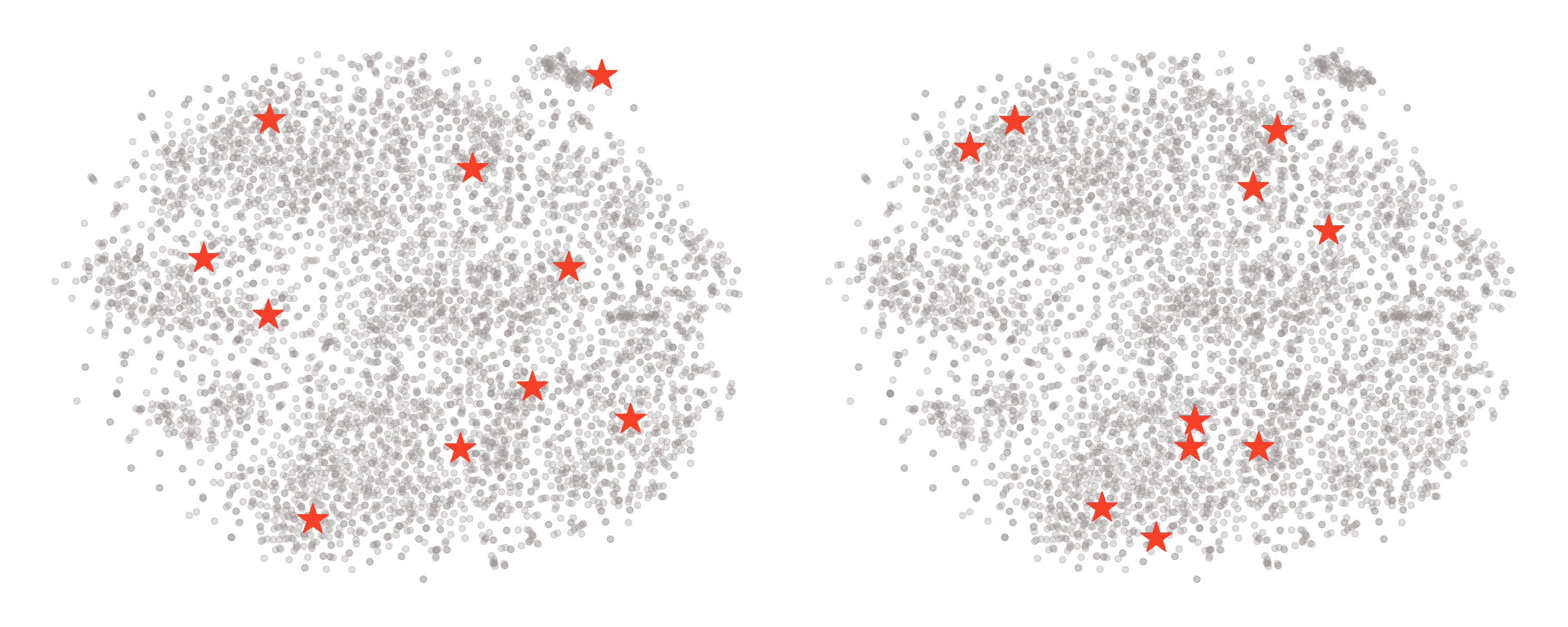}
        \put(35, 5){Ours}
        \put(85, 5){Random}
    \end{overpic}
    \caption{The sample distribution comparison on the final distilled images (marked as red stars) between our proposed DREAM\raisebox{0.09\baselineskip}{+} (left) and random sampling (right). }
    \label{fig:distribution}
\end{figure}

\begin{figure}[t]
\centering
\begin{subfigure}{0.24\textwidth}
    \includegraphics[width=\textwidth]{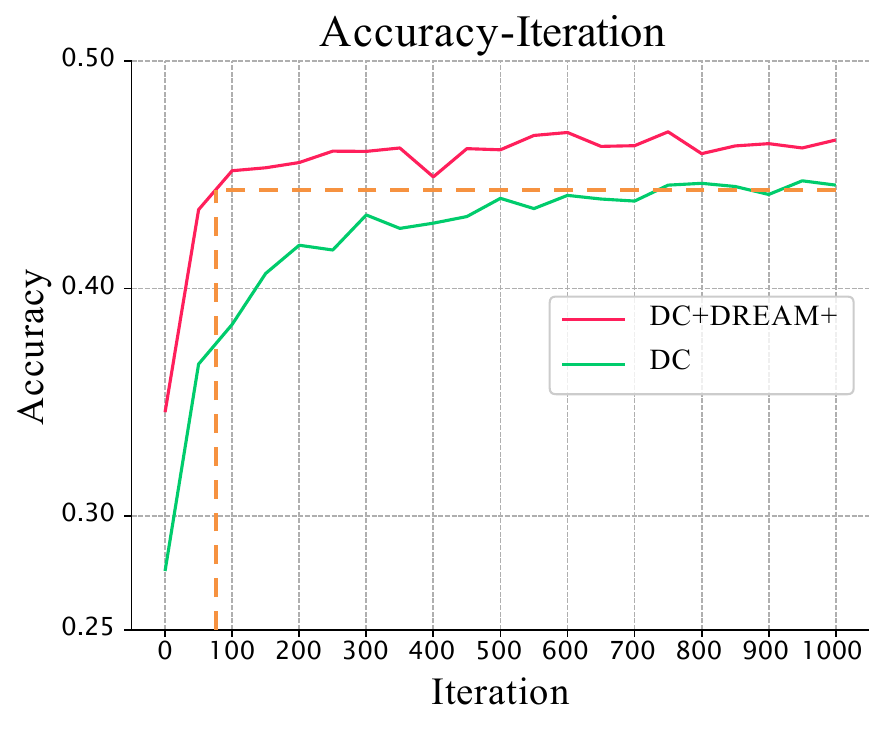}
    \caption{}
\end{subfigure}
\begin{subfigure}{0.24\textwidth}
    \includegraphics[width=\textwidth]{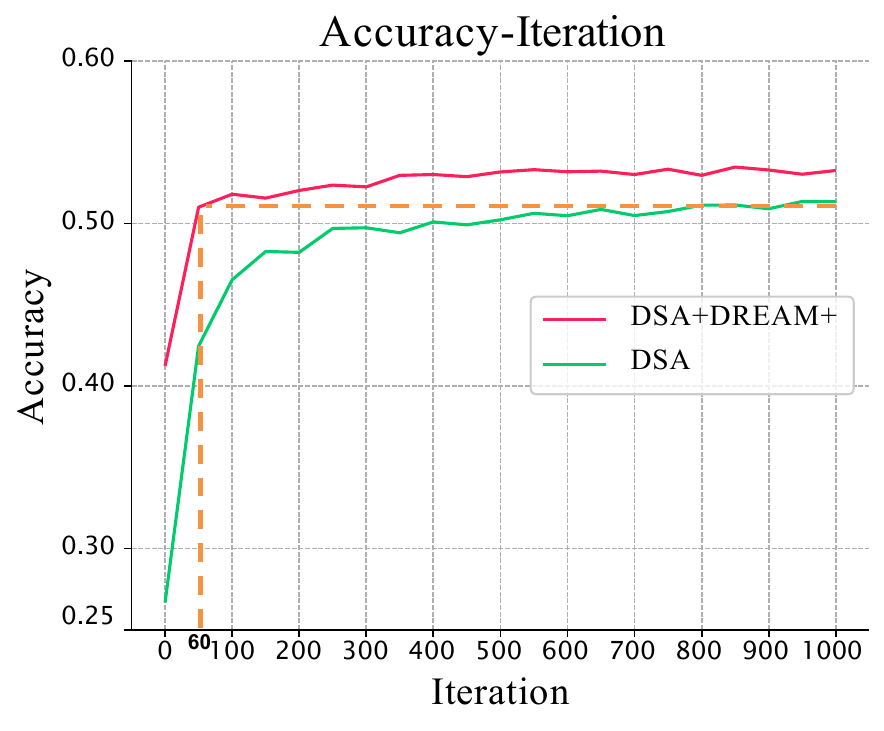}
    \caption{}
\end{subfigure}
\caption{Applying the DREAM\raisebox{0.09\baselineskip}{+} strategy brings stable performance and efficiency improvements for (a) DC and (b) DSA. }
\label{fig:curve}
\end{figure}

\begin{figure}
\centering
\includegraphics[width=\columnwidth]{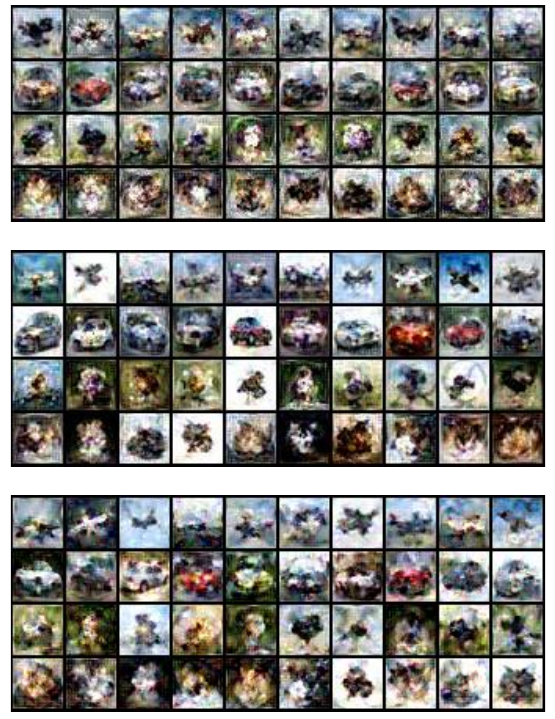}
\caption{Comparison of distilled datasets on CIFAR-10 (plane, car, dog, cat classes) for DC (top row), DC with DREAM strategy (middle row), and DC with DREAM\raisebox{0.09\baselineskip}{+} strategy (bottom row). On the basis of DREAM introducing more obvious categorical characteristics and diversity, DREAM\raisebox{0.09\baselineskip}{+} further adds diverse features to the synthesized images. Best viewed in color.}
\label{fig:dc-comp}
\end{figure}

\textbf{Sample Distribution Visualization.}
To more intuitively illustrate the efficacy of our DREAM\raisebox{0.09\baselineskip}{+} strategy in mimicing the original sample distribution, we employed t-SNE~\cite{van2008visualizing} visualization for both random sampling and DREAM\raisebox{0.09\baselineskip}{+}.
Referring to Figure~\ref{fig:distribution}, the t-SNE plot clearly illustrates the difference between the two methods. The results from DREAM\raisebox{0.09\baselineskip}{+} present a final distribution that evenly spans the entire category range. In contrast, random sampling can lead to significant bias in optimization results.
Furthermore, random sampling results show that most samples are drawn to the edge of the distribution. This observation highlights the bias introduced by boundary samples with larger gradients during the matching process.
By continuously providing appropriate gradient supervision and distribution supervision, DREAM\raisebox{0.09\baselineskip}{+} achieves more diverse and resilient distillation results.

\textbf{Appliance on More Methods.}
DREAM\raisebox{0.09\baselineskip}{+} is suitable for a variety of mainstream dataset distillation methods, including DC~\cite{zhao2020dc}, DSA~\cite{zhao2021dsa}, etc. We provide the training accuracy curve in Figure~\ref{fig:curve}.
Examining these curves carefully, we see that DREAM\raisebox{0.09\baselineskip}{+} requires only a fraction of the iterations to achieve the same performance compared to the original method. Specifically, in the case of DC and DSA, one-tenth of the number of iterations is sufficient to reach the original performance benchmark.
As training iterations increase, DREAM\raisebox{0.09\baselineskip}{+} continues to boost the performance.
All the above experiments are performed on CIFAR-10 with 10 IPC.

\textbf{Synthetic Image Visualization.}
In order to more intuitively understand the impact of DREAM\raisebox{0.09\baselineskip}{+} on distilled images, we visually compare the distillation results of DREAM\raisebox{0.09\baselineskip}{+}, DREAM and the baseline in Figure~\ref{fig:dc-comp}. 
First, images optimized with DREAM\raisebox{0.09\baselineskip}{+} and DREAM exhibit more distinct and obvious categorical characteristics, making them visually clear and easily identifiable. Second, DREAM\raisebox{0.09\baselineskip}{+} and DREAM also introduce greater diversity to distilled images, resulting in broader representation of the synthetic dataset. In addition, based on DREAM, DREAM\raisebox{0.09\baselineskip}{+} introduces more diverse feature representations. Clearer categorical characteristics, feature complexity, and higher image diversity work together to improve performance of synthetic datasets.

\begin{figure}
    \centering
    \includegraphics[width=0.3\textwidth]{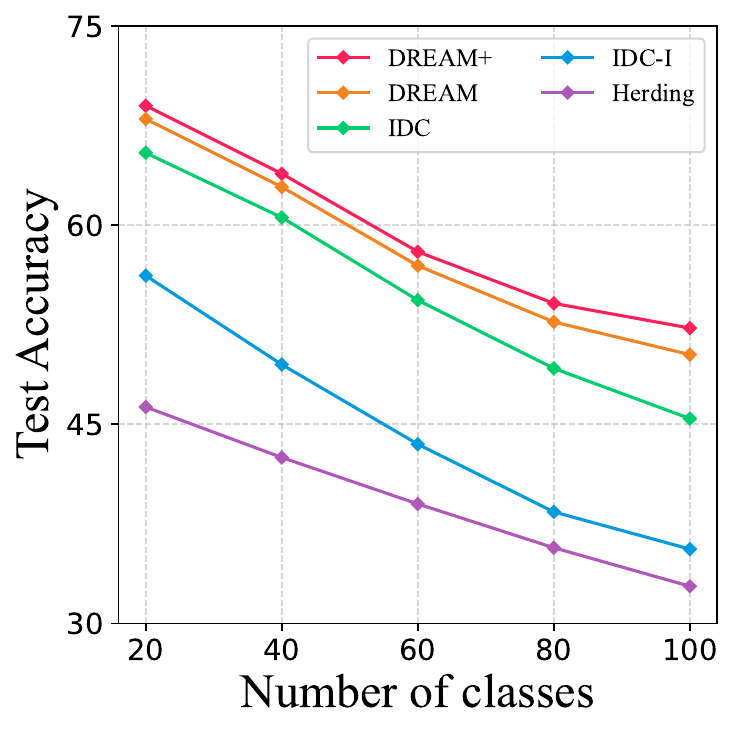}
    \caption{The continual learning accuracy curve.}
    \label{fig:continual}
\end{figure}

\subsection{Application on Continual Learning}
Dataset distillation is promising to apply in the continual learning~\cite{rebuffi2017icarl,aljundi2019gradient,wiewel2021condensed,idc}.
In Figure~\ref{fig:continual}, we evaluate the effectiveness of our proposed DREAM\raisebox{0.09\baselineskip}{+} strategy in the continual learning scenario. 
Following the experimental settings of~\cite{zhao2020dc,idc}, we conduct a 5-step class incremental experiment on CIFAR-100, in which 20 new classes were introduced at each step. 
We perform distillation synthesis on ConvNet-3 and verified it on ResNet-10.
Throughout the training process, DREAM\raisebox{0.09\baselineskip}{+} always maintains its performance advantage over other methods. Furthermore, the performance gap widens as the number of learning categories gradually increases. 
These results highlight the concept that improving the quality of distillation helps build clearer decision boundaries within the model, thereby better preserving discriminative information.

\section{Conclusion}
In this paper, we introduce a novel dataset distillation method named Dataset Distillation by Bidirectional Representative Matching (DREAM\raisebox{0.09\baselineskip}{+}). Our goal is to solve the training efficiency problem in dataset distillation. By sampling a representative set of original images for bidirectional matching, DREAM\raisebox{0.09\baselineskip}{+} further mitigates the instability of optimization, resulting in a more stable and robust training process.
DREAM\raisebox{0.09\baselineskip}{+} can be widely applied to existing dataset distillation frameworks, and significantly reduces the number of training iterations by more than 15 times without performance drop. 
This enhanced optimization stability contributes to superior final performance and improved generalization capabilities.
Furthermore, the improved efficiency of bidirectional matching opens the door to exploring more complex matching metrics in the future.

\begin{figure*}
\centering
\begin{subfigure}{0.28\textwidth}
    \includegraphics[width=\textwidth]{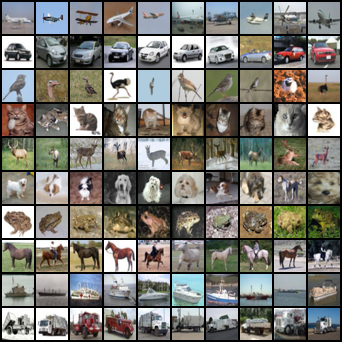}
    \caption{iteration=0}
\end{subfigure}
\begin{subfigure}{0.28\textwidth}
    \includegraphics[width=\textwidth]{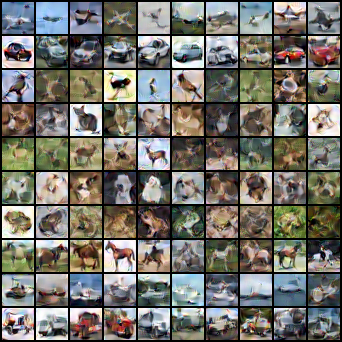}
    \caption{iteration=200}
\end{subfigure}
\begin{subfigure}{0.28\textwidth}
    \includegraphics[width=\textwidth]{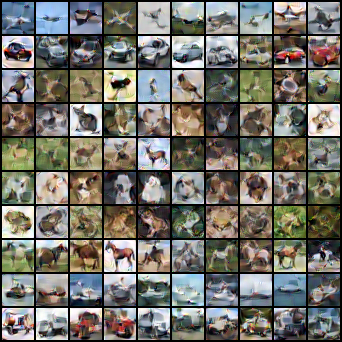}
    \caption{iteration=400}
\end{subfigure}
\begin{subfigure}{0.28\textwidth}
    \includegraphics[width=\textwidth]{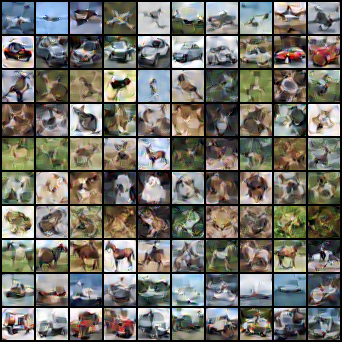}
    \caption{iteration=600}
\end{subfigure}
\begin{subfigure}{0.28\textwidth}
    \includegraphics[width=\textwidth]{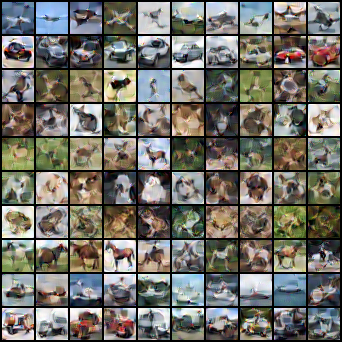}
    \caption{iteration=800}
\end{subfigure}
\begin{subfigure}{0.28\textwidth}
    \includegraphics[width=\textwidth]{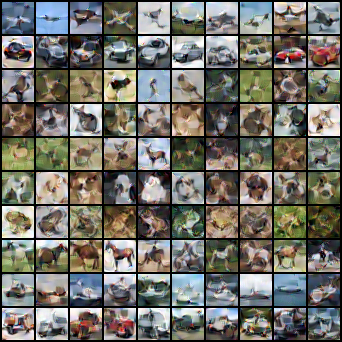}
    \caption{iteration=1000}
\end{subfigure}
\caption{Visualization of synthetic images at different training stages on CIFAR-10.}
\label{fig:image_iter}
\end{figure*}

\begin{figure*}
\centering
\begin{subfigure}{0.28\textwidth}
    \includegraphics[width=\textwidth]{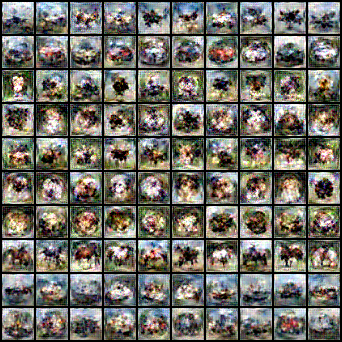}
    \caption{DC}
\end{subfigure}
\begin{subfigure}{0.28\textwidth}
    \includegraphics[width=\textwidth]{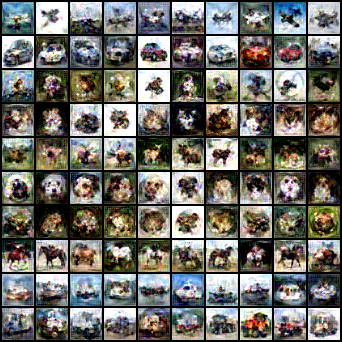}
    \caption{DC+DREAM}
\end{subfigure}
\begin{subfigure}{0.28\textwidth}
    \includegraphics[width=\textwidth]{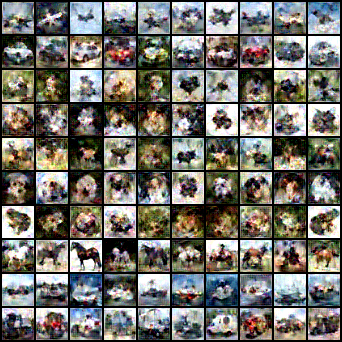}
    \caption{DC+DREAM\raisebox{0.09\baselineskip}{+}}
\end{subfigure}
\begin{subfigure}{0.28\textwidth}
    \includegraphics[width=\textwidth]{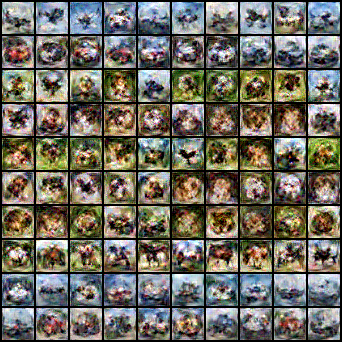}
    \caption{DSA}
\end{subfigure}
\begin{subfigure}{0.28\textwidth}
    \includegraphics[width=\textwidth]{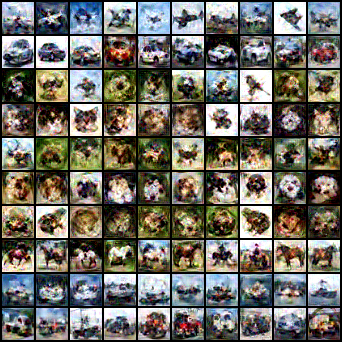}
    \caption{DSA+DREAM}
\end{subfigure}
\begin{subfigure}{0.28\textwidth}
    \includegraphics[width=\textwidth]{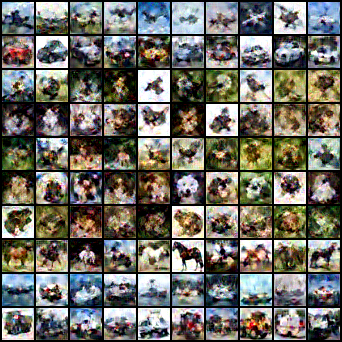}
    \caption{DSA+DREAM\raisebox{0.09\baselineskip}{+}}
\end{subfigure}
\caption{Applying DREAM enhances sample diversity, while DREAM\raisebox{0.09\baselineskip}{+} further improves image quality through feature alignment.}
\label{fig:sample}
\end{figure*}

\begin{figure*}
\centering
\begin{subfigure}{0.33\textwidth}
    \includegraphics[width=\textwidth]{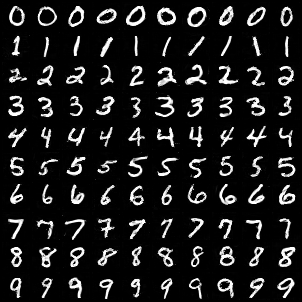}
    \caption{MNIST}
\end{subfigure}
\begin{subfigure}{0.33\textwidth}
    \includegraphics[width=\textwidth]{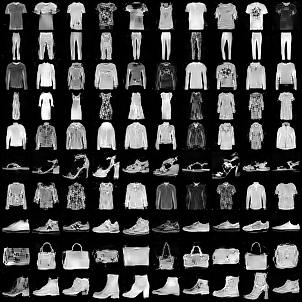}
    \caption{FashionMNIST}
\end{subfigure}
\begin{subfigure}{0.33\textwidth}
    \includegraphics[width=\textwidth]{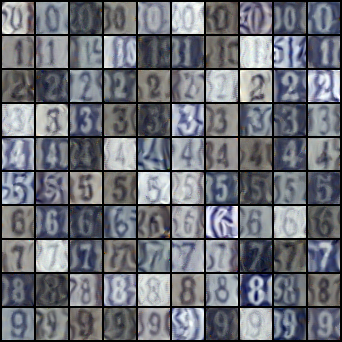}
    \caption{SVHN}
\end{subfigure}
\begin{subfigure}{0.32\textwidth}
    \includegraphics[width=\textwidth]{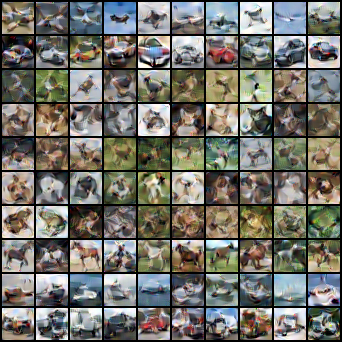}
    \caption{CIFAR-10}
    \label{fig:sample-CIFAR-10}
\end{subfigure}
\hskip 0.5em
\begin{subfigure}{0.32\textwidth}
    \includegraphics[width=\textwidth]{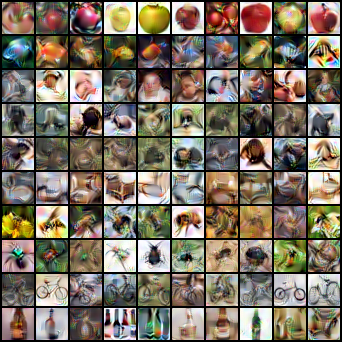}
    \caption{CIFAR-100}
    \label{fig:sample-CIFAR-100}
\end{subfigure}
\hskip 0.5em
\begin{subfigure}{0.32\textwidth}
    \includegraphics[width=\textwidth]{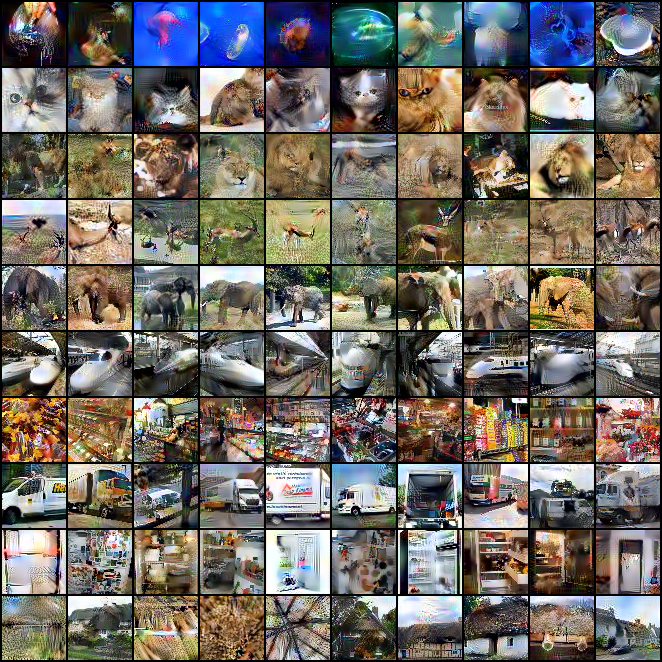}
    \caption{TinyImageNet}
    \label{fig:sample-tiny}
\end{subfigure}
\caption{Example visualizations of the distilled images on MNIST, FashionMNIST, SVHN, CIFAR-10, CIFAR-100 and TinyImageNet. }
\label{fig:sample-mnist}
\end{figure*}

\section{Limitations and Future Works}

Although our proposed DREAM\raisebox{0.09\baselineskip}{+} strategy greatly improves the training efficiency of optimization-based dataset distillation methods, it is worth noting that the computational requirements are still large, especially when dealing with larger image sizes and more classes.
Even with the efficiency enhancements introduced by DREAM\raisebox{0.09\baselineskip}{+}, these techniques may still encounter difficulties when processing very large datasets such as ImageNet~\cite{deng2009imagenet}. Furthermore, scaling up matching-based methods to accommodate more images per class may pose challenges.
Future research could focus on developing more computationally efficient distance measures or integrated core set methods to expand the number of images per class in image dataset distillation. These advancements could further enhance the scalability and practicality of dataset distillation for extensive and diverse image datasets.

\appendix[More Visualization Results]

\textbf{Visualization of Synthetic Image Variations. }
We performed a visual exploration of synthetic images evolving throughout the dataset distillation process, as shown in Figure~\ref{fig:image_iter}. This visual representation provides insight into the transformations the synthetic data undergoes during various training iterations. By observing changes in appearance, diversity, and alignment of these images, we can efficiently track convergence and evaluate the effectiveness of our proposed DREAM\raisebox{0.09\baselineskip}{+} strategy.
These visualizations not only provide a tangible sense of how the synthesis evolves, but also validate the stability and consistency of the bidirectional matching. Furthermore, they are strong evidence of DREAM\raisebox{0.09\baselineskip}{+}'s enhanced ability to generate high-quality synthetic data that faithfully captures the characteristics of the original dataset.

\textbf{Visualization of distillation dataset. } In order to more intuitively describe the impact on distilled images, we compared the dataset distillation results with and without using the DREAM\raisebox{0.09\baselineskip}{+} strategy and DREAM, as shown in Figure~\ref{fig:sample}. DREAM\raisebox{0.09\baselineskip}{+} enhances the distillation dataset from two different perspectives.
First, thanks to the newly introduced feature distribution matching, images optimized by DREAM\raisebox{0.09\baselineskip}{+} show more obvious classification characteristics.
Second, DREAM\raisebox{0.09\baselineskip}{+} introduces more diversity to distilled images. This diversification helps provide a richer representation in the dataset, which in turn improves the performance of distilled datasets.

We provide additional visualizations in Fig.~\ref{fig:sample-mnist}. Covering MNIST, FashionMNIST, SVHN, CIFAR-10, CIFAR-100, and TinyImageNet, these visualizations reiterate the advantages of DREAM\raisebox{0.09\baselineskip}{+} in various dataset distillation scenarios.

\ifCLASSOPTIONcompsoc
  \section*{Acknowledgments}
\else
  \section*{Acknowledgment}
\fi
This research is supported by the National Research Foundation, Singapore under its AI Singapore Programme (AISG Award No: AISG2-PhD-2021-08-
008). Yang You's research group is being sponsored by NUS startup grant (Presidential Young Professorship), Singapore MOE Tier-1 grant, ByteDance grant, ARCTIC grant, SMI grant and Alibaba grant.
The research is also supported by the National Natural Science Foundation of China (No. 62173302).

\ifCLASSOPTIONcaptionsoff
  \newpage
\fi

\normalem
\bibliographystyle{IEEEtran}
\bibliography{egbib}

\end{document}